\documentclass[10pt,twocolumn,letterpaper]{article}

\usepackage[pagenumbers]{iccv} 

\definecolor{iccvblue}{rgb}{0.21,0.49,0.74}
\usepackage[pagebackref,breaklinks,colorlinks,allcolors=iccvblue]{hyperref}
\usepackage{multirow}
\usepackage{float}
\def\paperID{7607} 
\def\confName{ICCV}
\def\confYear{2025}

\title{TPC: Cross-Temporal Prediction Connection for Vision-Language Model
Hallucination Reduction}

\author{Chao Wang\\
Shanghai University\\
{\tt\small cwang@shu.edu.cn}
\and
Weiwei Fu\\
Shanghai University\\
{\tt\small fuweiwei2001@shu.edu.cn}
\and
Yang Zhou\\
Shanghai University\\
{\tt\small saber\_mio@shu.edu.cn}
}
\begin{document}
\maketitle
\begin{abstract}
Vision-language models (VLMs) have achieved remarkable advancements, capitalizing on the impressive capabilities of large language models (LLMs) across diverse tasks. Despite this, a critical challenge known as hallucination occurs when models overconfidently describe objects or attributes absent from the image, a problem exacerbated by the tendency of VLMs to rely on linguistic priors. This limitation reduces model reliability in high-stakes applications. In this work, we have observed the characteristic of \textbf{logits' continuity consistency enhancement} and introduce a straightforward and efficient method, Cross-\textbf{T}emporal \textbf{P}rediction \textbf{C}onnection (\textbf{TPC}), designed to enhance the semantic consistency of logits by connecting them temporally across timesteps. TPC amplifies information flow and improves coherence, effectively reducing hallucination. Extensive experiments show that TPC surpasses existing representatives, delivering superior performance in both accuracy and efficiency while maintaining robustness in open-ended text generation tasks.
\end{abstract}    
\section{Introduction}
\label{sec:intro}
In the field of vision-language models (VLMs), integrating a visual encoder with a large language model (LLM) and learning a projector layer is a common approach to impart visual capability \cite{liu2023llava,chen2024internvl}. VLMs have made substantial progress by capitalizing on the strong performance of LLMs across a wide range of tasks \cite{li2020oscar,10445007,touvron2023llama,raffel2020exploring,zhou2022learning}. However, a persistent challenge remains in generating descriptions that are both grammatically fluent and accurately reflective of visual content. This issue, often referred to as the hallucination occurs when the model confidently generates descriptions of objects or attributes that are not actually present in the corresponding image \cite{gunjal2024detecting,guan2024hallusionbench,leng2024mitigating,zhou2023analyzing}. Hallucination tends to accompany the introduction of LLMs, largely due to VLMs’ over-reliance on linguistic priors \cite{zhang2024vision,gupta2022swapmix,wu2023role,hamidieh2024identifying,zhang2022counterfactually}, which can compromise model reliability in applications where precise visual understanding is critical, such as autonomous driving systems or medical image interpretation \cite{dai2024vistarag,bhadra2021hallucinations,cohen2018distribution}. 

\begin{figure}
    \centering
    \includegraphics[width=\linewidth]{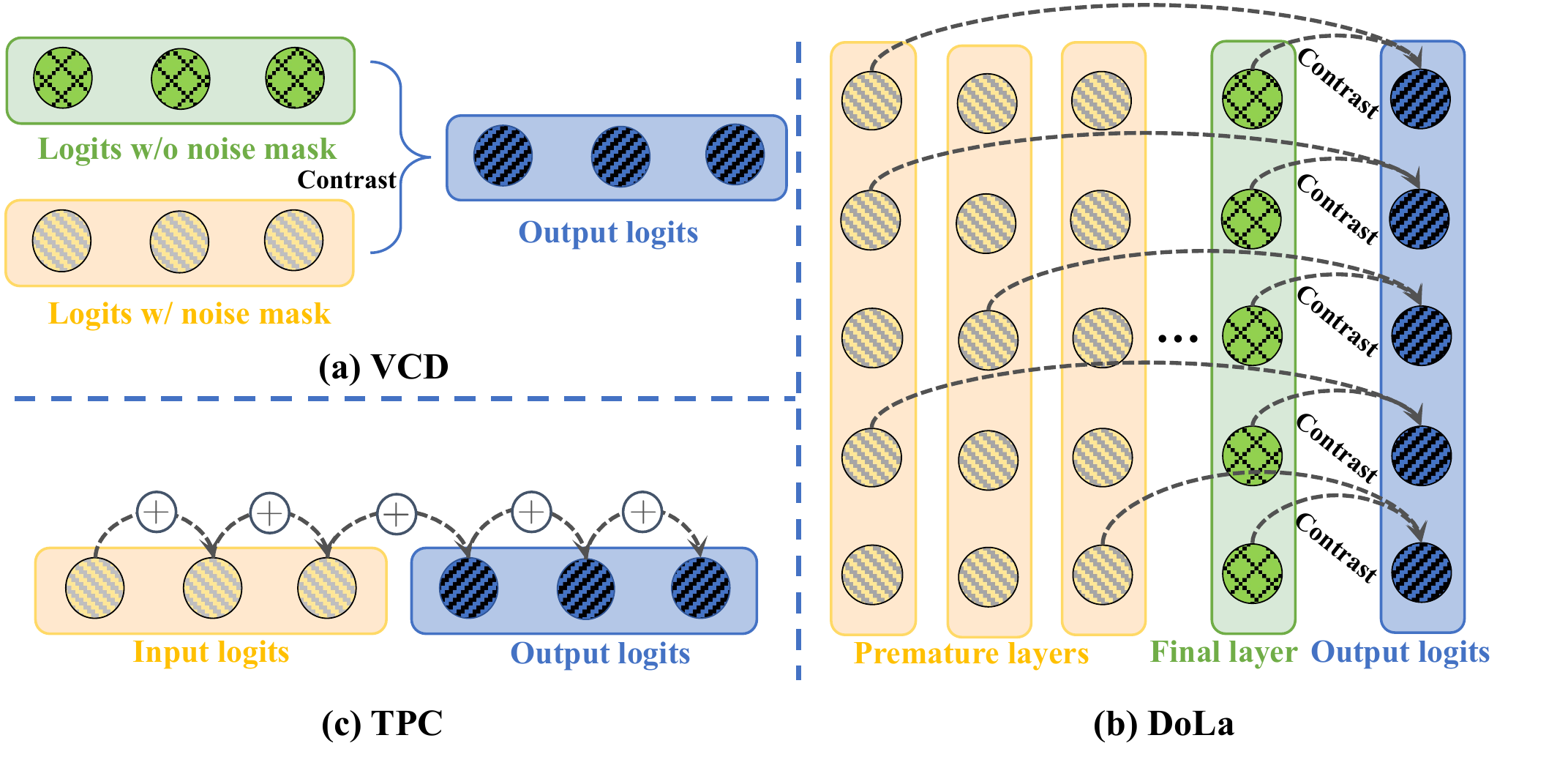}
    \caption{(a) VCD contrasts the normal logits with those obtained from inputting a noise-masked image. (b) DoLa compares each token's logits with the early-exit logits from other layers, selecting the logits with the largest Jensen-Shannon divergence for contrastive decoding. (c) Our TPC connects the logits of adjacent time steps without the need to contrast them with other logits.}
    \label{fig:baselines}
\end{figure}

Recent researches \cite{rohrbach2018object,leng2024mitigating,huang2024opera,favero2024multi,wu2024logical,rawte2023survey,jiang2024hallucination,bai2024hallucination} have primarily focused on object hallucination, which can be categorized into three types: category, attribute, and relational hallucinations. These hallucinations arise from both the biased distribution of training data and an imbalance between the capacities of the visual encoder and the LLM decoder. In most VLMs, the LLM has significantly more parameters than the visual encoder, leading to greater attention weights being assigned to text tokens, with visual tokens receiving minimal attention \cite{liu2024paying,diao2024EVE}. This disparity is a key reason why VLMs are more inclined to rely on linguistic priors.

Various approaches have analyzed hallucination causes from the perspective above and introduced new mitigation strategies. Some methods propose new datasets for instruction tuning \cite{hu2023ciem,liu2023mitigating,chen2024internvl} or refine human feedback preference tuning \cite{sun2023aligning,favero2024multi}, which, while effective, are often inefficient. Consequently, recent post-hoc methods have gained traction, focusing on dimensions such as linguistic priors and attention weights to design training-free, flexible, and efficient post-hoc strategies \cite{chen2024context,yin2023woodpecker,yue2024less}. VCD \cite{leng2024mitigating} is a typical representative: as shown in Figure \ref{fig:baselines} (a), VCD generates two logits at each time step—one from standard VLM generation and another from the image masked with Gaussian noise. By contrastively decoding these logits, VCD reduces the influence of linguistic priors. Such contrastive decoding has led to various improved versions \cite{wang2024mitigating,zhang2024debiasing}. However, a clear limitation is that these methods double the inference time by generating logits twice per time step. DoLa \cite{chuang2023dola} also employs contrastive decoding but compares the logits at each time step with early exit\footnote{Early exit refers to directly feeding the model’s internal hidden states into the LLM head, producing an output of the same size as the logits.} to filter noise from early model layers, allowing the model to focus more on factual knowledge, as illustrated in Figure \ref{fig:baselines} (b).

Unlike filtering out the early exits from logits, we hypothesize that logits contain untapped information and knowledge. Since factual information can be extracted from each logit through the network’s depth, could connecting logits temporally across time steps enhance the information within logits? Based on this, we propose a Cross-Temporal Prediction Connection (TPC) method, as illustrated in Figure \ref{fig:baselines} (c). We conduct extensive experiments and theoretical derivations showing that simply connecting logits across different time steps can strengthen their semantic consistency, integrate information from preceding time steps, and notably improve long-text generation quality, outperforming VCD and DoLa in robustness, efficiency, and overall performance. Our main \textbf{contributions} are as follows:
\begin{enumerate}
\item We propose TPC, a straightforward, training-free, highly effective, and plug-and-play multimodal hallucination mitigation strategy.
\item We discover the logits' continuity consistency enhancement characteristic and provide theoretical and experimental validation, offering a robust interpretative basis for its effectiveness.
\item Through extensive experimentation, we demonstrate that TPC significantly alleviates object hallucination, excels in open-ended text generation tasks, and shows strong resilience against hallucinated tokens in the input, surpassing the performance of representative contrastive decoding approaches.
\end{enumerate}

\section{Related Work}
\label{sec:rel}
\paragraph{Hallucination in VLMs.} 
Hallucinations in VLMs primarily stem from two sources \cite{bai2024hallucination,zhou2023analyzing}. First, there is a lack of high-quality image-text data with sufficient diversity \cite{liu2023aligning,yu2024hallucidoctor}. These datasets often exhibit statistical biases \cite{deletang2023language,li2023evaluating}, which models can easily internalize, leading to biased responses that compromise fairness. Second, due to weak alignment between modalities, the LLM allocates the majority of attention weights to language tokens over visual tokens \cite{guan2024hallusionbench,rohrbach2018object,wang2023llm}. Consequently, the LLM dominates and VLMs become overly reliant on linguistic priors. Research on hallucinations in VLMs mainly focuses on modality consistency, examining whether the generated text description aligns with the image content \cite{jiang2024hallucination}. More specifically, most multimodal hallucination studies address object hallucination, which can be divided into three types according to the literature: category, attribute, and relational hallucinations. These focus on the consistency of the model’s descriptions with the object categories, color and shape attributes, and positional relationships depicted in the images \cite{bai2024hallucination}.

\paragraph{Post-Hoc for Hallucination Mitigation.} 
In contrast to approaches that rely on retraining, post-hoc methods aim to deliver effective hallucination mitigation without the need for additional training. OPERA \cite{huang2024opera}, for instance, observes the behavior of hallucinated tokens within attention layers and introduces a backtracking mechanism to avoid generating these tokens. LogicCheckGPT \cite{wu2024logical} uses only prompts to assess the consistency between text and images, helping detect hallucinations. Among recent works, contrastive decoding (CD) \cite{li2022contrastive} methods stand out, they induce hallucinated tokens through various means and then compare the logits of the model’s normal output with those of the hallucinated tokens (termed “hallucination logits”), filtering out hallucinated components in the logits. For example, VCD \cite{leng2024mitigating} uses Gaussian noise masking on the image to obtain hallucination logits, ICD \cite{wang2024mitigating} generates hallucination logits directly via specific prompts, and RVD \cite{zhong2024investigating} contrasts logits from inputs with and without hallucinations. While these contrastive methods avoid the need for additional training and dataset collection, they significantly reduce efficiency by doubling inference time, as two generations are required per time step. DoLa \cite{chuang2023dola} is another contrastive decoding approach but avoids the need for two token generations by comparing the final output logits with early exit from a premature layer. However, this requires additional memory due to the need for premature layer search. Our TPC avoids both of these limitations, providing a nearly “free lunch” with superior performance and robustness.

\section{Method}
\label{sec:method}

\begin{figure}[htbp]
    \centering
    \includegraphics[width=0.85\columnwidth]{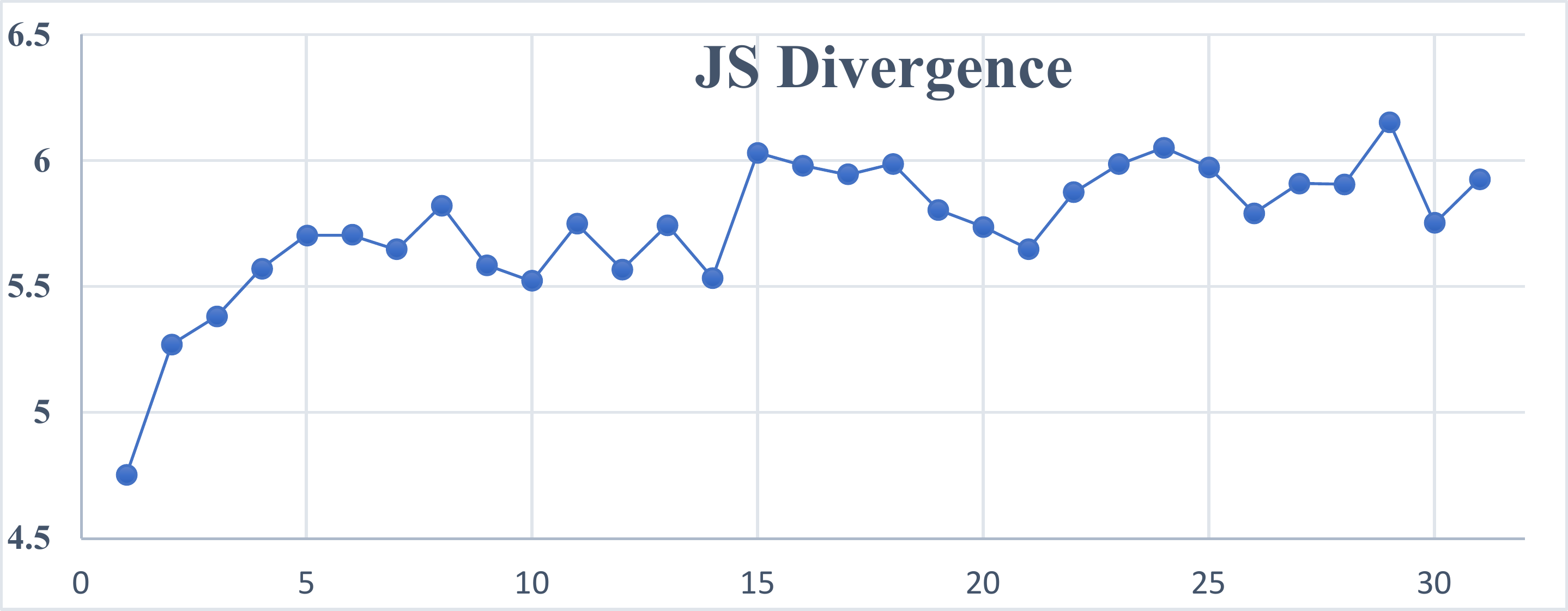}
    \caption{The relationship between the JS divergence of logits and the distance. The $x$-axis represents the distance corresponding to the time step of the logits. The divergence of each point is obtained by calculating and averaging the logits of the output sequences of LLaVA in pairs.  }
    \label{fig:js-kl}
\end{figure}

\begin{figure}[htbp]
    \centering
    \begin{subfigure}{0.85\columnwidth}
        \centering
        \includegraphics[width=\columnwidth]{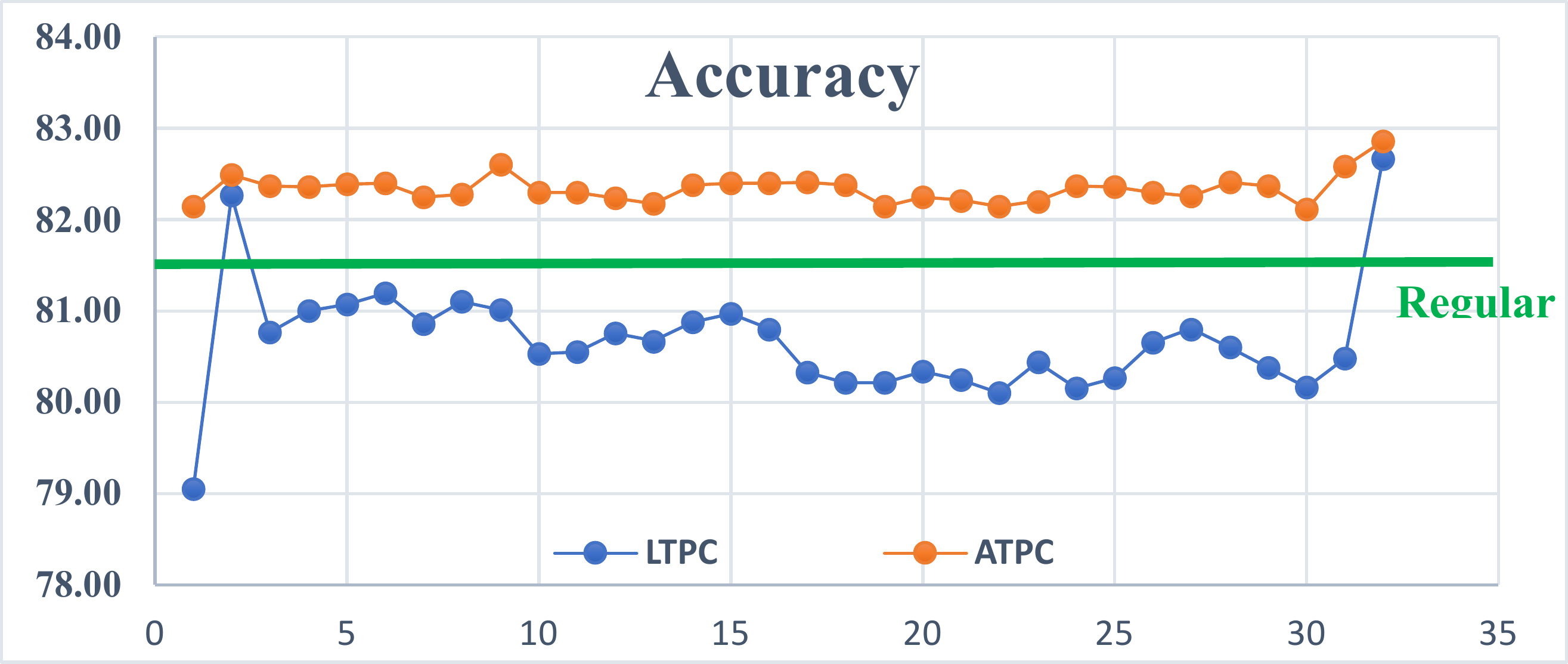}
    \end{subfigure}
    
    \vspace{0.2cm} 
    
    \begin{subfigure}{0.85\columnwidth}
        \centering
        \includegraphics[width=\columnwidth]{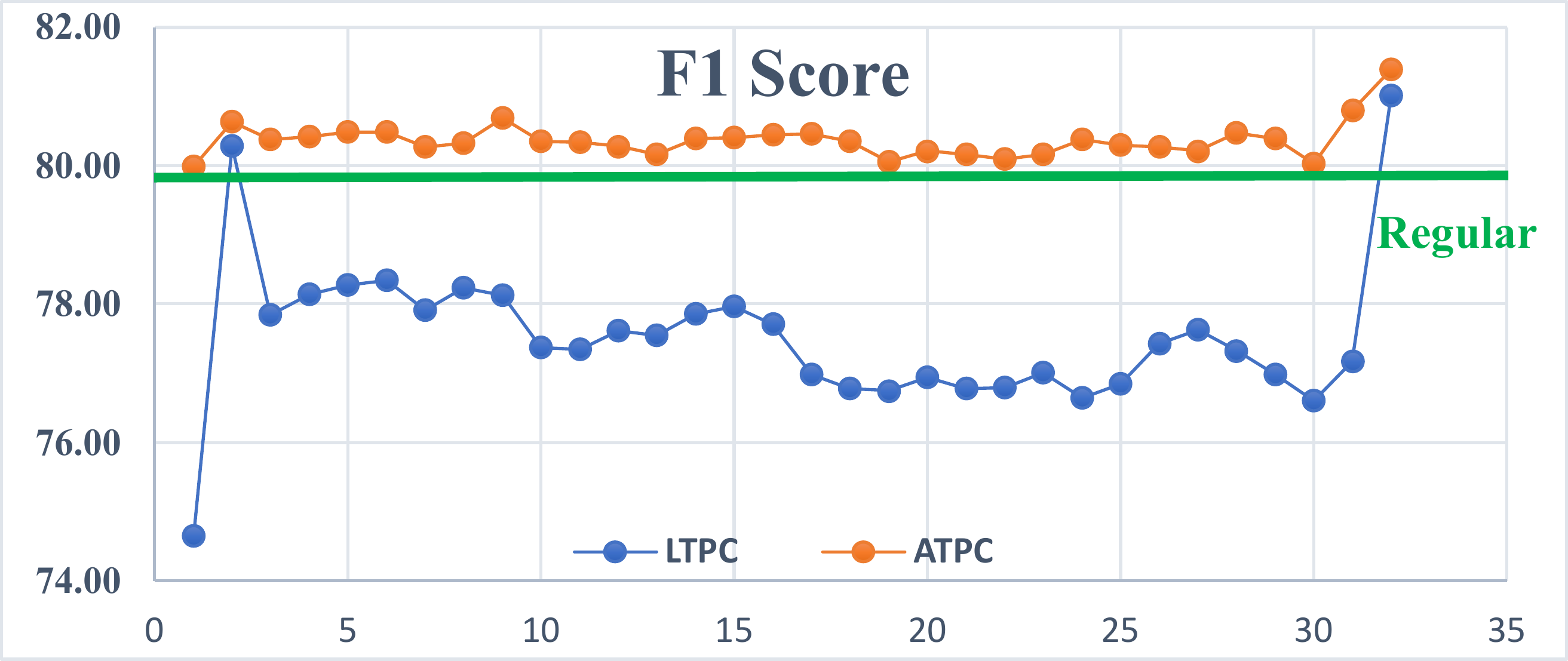}
    \end{subfigure}
    \caption{A sliding window is used to validate the two different connection strategies. The $x$- axis represents the position of the segmented windows in the sequence. The further to the right, the closer the position of the window is to the logit at the current time step. The upper part of the figure shows the accuracy of the two strategies, while the lower part displays the corresponding F1 score. Regular indicates the scores from the nucleus sampling.}
    \label{fig:sliding window}
\end{figure}

\begin{figure*}[t]
    \centering
    \includegraphics[width=0.95\linewidth]{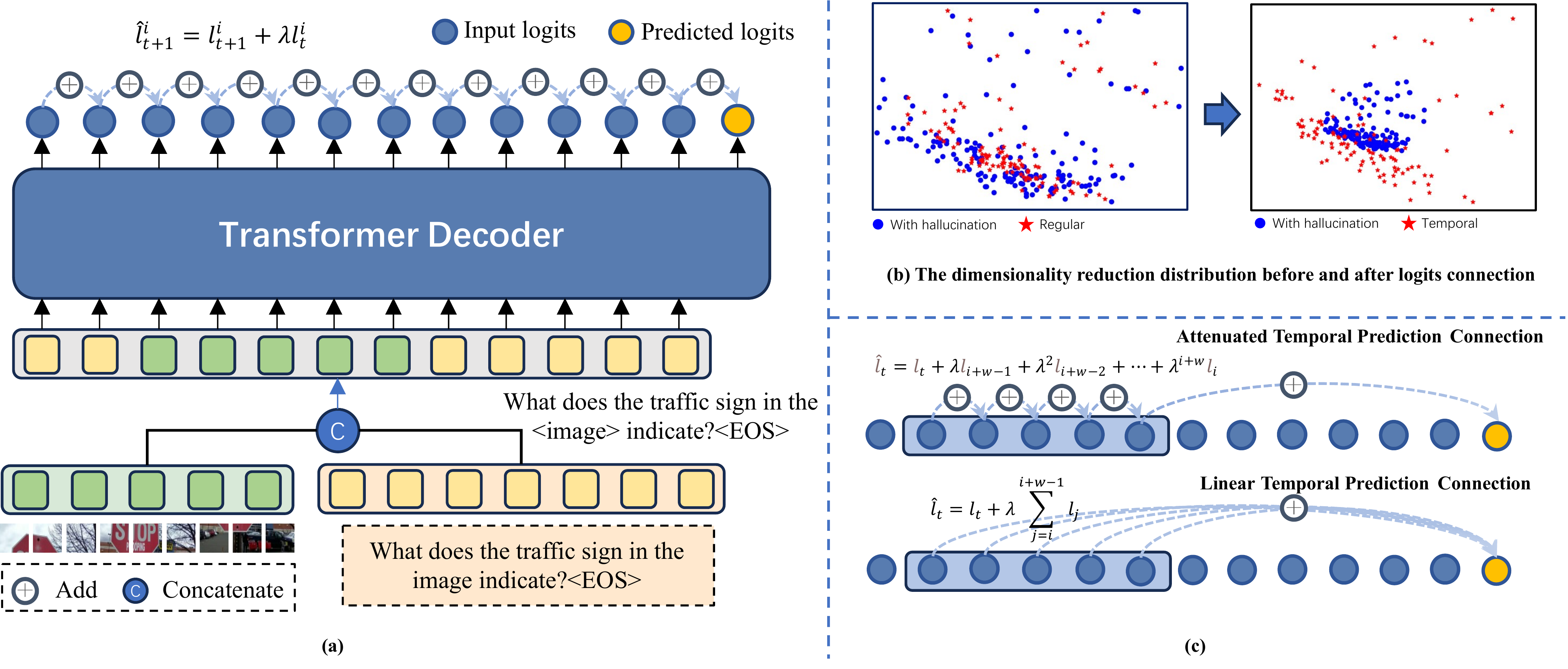}
    \caption{\textbf{An illustration and overview of TPC.} (a) Sequentially connecting logits across different time steps enhances and propagates temporal information. (b) Dimensionality reduction maps the logits of hallucinated tokens, regular tokens, and TPC tokens into a two-dimensional space, where TPC displays a more dispersed distribution, separated from the hallucinated tokens. (c) ATPC and LTPC.}
    \label{fig:method main}
\end{figure*}

\subsection{Preliminaries}
\label{sec:preli}
The dominant architecture for current vision-language models primarily consists of a visual encoder $e(\cdot)$ paired with a decoder-only large language model $Decoder_{LLM}$. As shown in Figure \ref{fig:method main} (a), the visual encoder processes input images $x$ and transforms them into image tokens, which are projected into the language embedding space through a projector layer. These image embeddings are then fed into the decoder layers of the language model, where they are processed alongside text embeddings. Each decoder layer in the $Decoder_{LLM}$ incrementally refines the representations, and after passing through all layers, the final hidden states are obtained. These hidden states are then mapped to the language model's vocabulary space through the LLM head $\phi (\cdot)$ to predict the next token in the sequence:
\begin{equation}
    P(y_1, y_2, \dots, y_T | x) = \prod_{t=1}^{T} P(y_t | y_1, y_2, \dots, y_{t-1}, x),
\end{equation}
\begin{equation}
    l_t = \phi(Decoder(h_{t-1}, \dots, h_1, e(x)))
\end{equation}
$x$ and $y$ represent the input image-text pairs and the output tokens of the model respectively. $h$ refers to the input hidden state, and $l$ is the logit.

Recent works \cite{leng2024mitigating, chuang2023dola} show that logits are very likely to contain a lot of hallucinatory and even harmful information, and different contrastive methods are used to filter out such information from logits. The method in this work also starts from logits and yields interesting observations. 
In the following sections, we will describe the observed characteristics of information consistency and complementarity between adjacent logits.

\subsection{Logits' Continuity Consistency Enhancement}

When analyzing the output logits of large language models (LLMs), we find that the distribution difference between adjacent logits is significantly lower than that between non-adjacent logits. Generally, there is a pattern that the distribution difference between logits increases as the interval between them becomes larger. As shown in Figure \ref{fig:js-kl}, we use the Jensen-Shannon (JS) divergence to measure the difference between two distributions. Overall, the JS divergence increases as the distance between two logits increases. Therefore, we speculate that although the tokens with the highest probability corresponding to logits at different time steps vary greatly, the local distributions of logits at adjacent time steps are similar. These similar local distributions represent the internal correlation and consistency of neighboring logits. To further explore the internal relationship between adjacent logits, we have designed two ways of connecting logits, namely Linear Temporal Prediction Connection (LTPC) and Attenuation Temporal Prediction Connection (ATPC). 

\paragraph{Linear Temporal Prediction Connection.} 
Initially, a fixed-size sliding window is considered, defined as \( T_i^w = \{l_i, l_{i+1}, ..., l_{i+w-1}\} \), which includes the logits of consecutive tokens in the sequence. As shown in Figure \ref{fig:method main} (c), the logits within this window are multiplied by a weighting factor \(\lambda\) and added to the logits at the next timestep:
\begin{equation}
    \hat{l}_t = l_t + \lambda \sum_{j=i}^{i+w-1} l_j
\end{equation}
We believe that the logits of a continuous sequence contain rich contextual information, which can enhance the scope and consistency of generated logits. We use LLaVA1.5-7B to explore the impact of windows at different time steps on the logit generated at the current time step. Performance is evaluated on POPE MSCOCO \cite{li2023evaluating}. The logits obtained from the input sequence are divided into 32 segments, with each segment serving as a window. Each window contains 20 logits. Each time, we connect the logits within one window to the currently generated logits in the LTPC manner, and evaluate the Accuracy and F1 score. As shown in Figure \ref{fig:sliding window}, \textbf{Regular} denotes nucleus sampling, and connecting the logits from the initial 20 tokens of input sequence results in a noticeable drop in accuracy and F1 score compared to Regular. This may be because the initial system prompt in the input does not contribute substantively to answering image-based questions. During the window movement, performance fluctuates significantly and remains below that of Regular. However, when the window closest to the next time step is connected, performance rises sharply. Due to semantic gaps and missing information in discontinuous logits, the final 20 logits of the input sequence maintain complete consistency with the logits generated at the next time step. This results in a rapid increase in accuracy. We refer to this improvement through connecting continuous logits as \textit{logits' continuity consistency enhancement}.

\paragraph{Attenuation Temporal Prediction Connection.} To further optimize the continuity enhancement effect, we propose an attenuation connection strategy based on the observations above. As shown in Figure \ref{fig:method main} (c), to maintain consistency among adjacent logits, we do not connect all logits within the window directly to the final time step’s logits. Instead, each time step’s logits are sequentially added to the next adjacent time step, as formulated below:
\[
   \hat{l}_t = l_t + \lambda \hat{l}_{i+w-1},
\]
\begin{equation}
    \hat{l}_{i+w-1} = l_{i+w-1} + \lambda \hat{l}_{i+w-2}
\end{equation}
where \(\lambda\) is the attenuation coefficient, controlling each preceding time step’s contribution to the current time step’s logit. Thus, the final expression becomes:
\begin{equation}
    \hat{l}_t = l_t + \lambda l_{i+w-1} + \lambda^2 l_{i+w-2} + \dots + \lambda^{i+w} l_i
\end{equation}
Through gradual attenuation, logits closer to the current time step receive higher weights, while those further away diminish in influence, thereby reinforcing the logits' continuity consistency enhancement effect during generation. This attenuation mechanism ensures that the model prioritizes high-value information from nearby time steps at each generation step while reducing noise potentially introduced by distant information. We conduct the sliding window experiment again, and as shown in Figure \ref{fig:sliding window}, the model's performance consistently exceeds that of Regular and LTPC as the window moves, with more stable variations. This result continues to demonstrate the \textit{logits' continuity consistency enhancement} characteristic, aligning with our hypothesis.

\begin{figure*}[htbp]
  \centering
  \begin{subfigure}{0.32\linewidth}
    \includegraphics[width=1.0\linewidth]{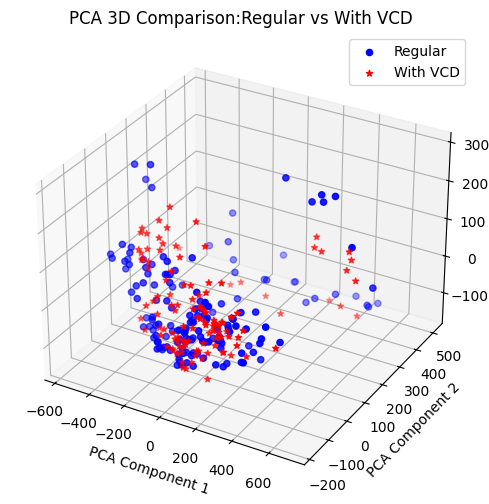}
    \caption{VCD vs Regular.}
    \label{fig:PCA_vcd}
  \end{subfigure}
  \hfill
  \begin{subfigure}{0.32\linewidth}
    \includegraphics[width=1.0\linewidth]{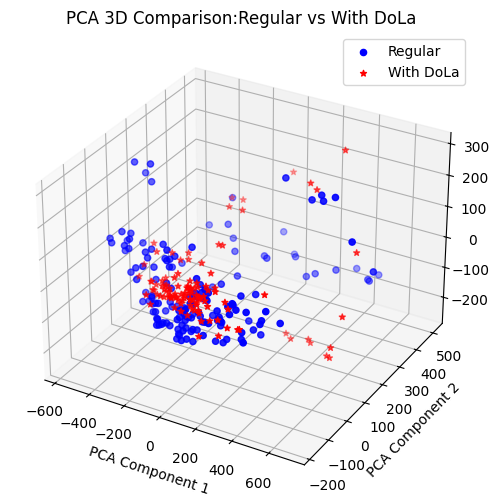}
    \caption{DoLa vs Regular.}
    \label{fig:PCD_dola}
  \end{subfigure}
  \hfill
  \begin{subfigure}{0.32\linewidth}
    \includegraphics[width=1.0\linewidth]{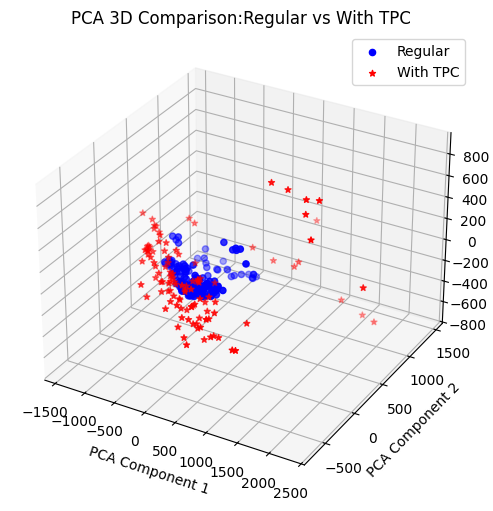}
    \caption{TPC vs Regular.}
    \label{fig:short-c}
  \end{subfigure}
  \caption{PCA analysis maps the model’s output logits into 3D space, showing a clear separation between TPC and Regular distributions.}
  \label{fig:PCA_time}
\end{figure*}

\subsection{Logits Distribution Analysis}

When generating answers, fluctuations in the contextual semantics of the input may lead the model to overly depend on certain tokens. By connecting logits over time, each timestep’s logits integrate input from other relevant timesteps, resulting in more diversified information and improved consistency across the sequence. 
We apply principal component analysis (PCA \cite{abdi2010principal}) to reduce the dimensionality of logits from the model’s output sentences to observe their distribution, with each point representing an output token. In Figure \ref{fig:method main} (b), we project the logits corresponding to hallucinated tokens, non-hallucinated tokens, and TPC-generated logits into two-dimensional space. The TPC distribution shows a separation from the hallucinated token distribution. More specifically, in Figure \ref{fig:PCA_time} (c), we map the logits generated by Regular, VCD, and DoLa into three-dimensional space. The distributions for Regular, VCD, and DoLa show no significant differences, while TPC’s more dispersed distribution indicates increased sentence-level variance and enhanced representational capacity, further validated in the experimental section with open-ended generation tasks.

\section{Experiments}
\subsection{Implementation details}
\paragraph{Datasets.} 
POPE~\cite{li2023evaluating} is a recent dataset focused on object hallucination, divided into three splits: Random, Popular, and Adversarial. In Random, objects absent from the image are chosen randomly. Popular selects high-frequency objects, while Adversarial prioritizes objects that typically co-occur with those in the image but are currently missing. The images in POPE are sourced from MSCOCO, A-OKVQA, and GQA. MME~\cite{fu2023mme} is divided into two categories that assess the model's Perception and Cognition tasks. These two tasks are composed of various domain-specific datasets and provide a comprehensive score. Unlike the first two, which only require Yes/No answers, MMHal-Bench~\cite{sun2023aligning} evaluates the model's ability to generate open-ended responses, focusing on whether the generated text contains sufficient ground-truth information and avoids hallucination.

\begin{table*}[htbp]
\centering
\resizebox{\textwidth}{!}{%
\begin{tabular}{|cccccccccccccccccccc}
\hline
\multicolumn{20}{|c|}{\cellcolor[HTML]{C0C0C0}POPE MSCOCO}                                                                                                                                                                                                                                                                                                                                                                                                                                                                                                                                                                                                                                                                                                                                                                                                                                                                            \\ \hline
                                                    & \multicolumn{1}{c|}{}                                     &                       & \multicolumn{4}{c}{Popular}                                                                                                                                                            &                       & \multicolumn{4}{c}{Random}                                                                                                                                                             &                       & \multicolumn{4}{c}{Adversarial}                                                                                                                                                        & \multicolumn{1}{c|}{} &                                                          & \multicolumn{1}{c|}{}                                                         \\
\hline
\end{tabular}%
}
\caption{Results on the POPE full set with LLaVA1.5-7\&13b and QwenVL, where baseline results are obtained from our experimental implementation. Bold font indicates the
best result. The results for the baselines are reproduced based on the original papers.}
\label{POPE All}
\end{table*}

\begin{table}[]
\centering
\Large
\resizebox{0.85\columnwidth}{!}{%
\begin{tabular}{|cc|ccc|}
\hline
\multirow{2}{*}{Method}                       & \multirow{2}{*}{Decoding} & \multicolumn{3}{c|}{MiniGPT4}                             \\
                                              &                           & CHAIRs $\downarrow$ & CHAIRi $\downarrow$ & Accuracy $\uparrow$ \\ \hline
\multirow{2}{*}{PAI}                          & Greedy                    & 27.0              & 9.1               & 76.53             \\
                                              & Beam Search               & 22.8              & 8.5               & 76.62             \\ \hline
\multirow{2}{*}{TPC}                          & Greedy                    & 30.4              & 10.1              & \textbf{76.96}    \\
                                              & Beam Search               & 28.4              & 9.7               & \textbf{76.87}    \\ \hline
\multicolumn{1}{|l}{\multirow{2}{*}{PAI+TPC}} & Greedy                    & \textbf{24.2}     & \textbf{8.2}      & \textbf{76.96}    \\
\multicolumn{1}{|l}{}                         & Beam Search               & \textbf{22.6}     & \textbf{8.1}      & \textbf{76.87}    \\ \hline
\end{tabular}%
}
\caption{\textbf{Comparison with PAI.} We combine different decoding strategies and compare CHAIR on MSCOCO and Accuracy on POPE. When using MiniGPT4, combining PAI with TPC can achieve the best results. }
\label{minigpt}
\end{table}

\begin{table*}[htbp]
\centering
\resizebox{\textwidth}{!}{%
\begin{tabular}{|c|cccccccccc|c|}
\hline
Method       & \multicolumn{1}{l}{Existence} & \multicolumn{1}{l}{Count} & \multicolumn{1}{l}{Position} & \multicolumn{1}{l}{Color} & \multicolumn{1}{l}{Posters} & \multicolumn{1}{l}{Celebrity} & \multicolumn{1}{l}{Scene} & \multicolumn{1}{l}{Artwork} & \multicolumn{1}{l}{OCR} & \multicolumn{1}{l|}{Landmark} & \textbf{Total score} \\ \hline
Regular      & $182.33_{\pm6.15}$            & $122.83_{\pm11.03}$       & $120.50_{\pm6.76}$           & $146.17_{\pm9.20}$        & $128.74_{\pm4.86}$          & $98.97_{\pm29.60}$            & $147.25_{\pm3.19}$        & $106.43_{\pm3.36}$          & $103.25_{\pm13.49}$     & $131.20_{\pm5.07}$            & $1297.67_{\pm33.47}$ \\
VCD          & $181.16_{\pm3.85}$            & $127.17_{\pm13.24}$       & $129.17_{\pm11.53}$          & $154.33_{\pm10.19}$       & $129.81_{\pm8.33}$          & $121.53_{\pm6.57}$            & $151.15_{\pm2.51}$        & $107.10_{\pm3.88}$          & $111.25_{\pm9.52}$      & $138.13_{\pm4.41}$            & $1352.32_{\pm38.60}$ \\
DoLa         & $188.67_{\pm4.77}$            & $125.33_{\pm11.67}$       & $126.33_{\pm7.89}$           & $158.17_{\pm12.97}$       & $130.27_{\pm5.35}$          & $113.32_{\pm4.79}$            & $151.55_{\pm3.07}$        & $106.85_{\pm3.46}$          & $111.75_{\pm15.55}$     & $132.88_{\pm5.13}$            & $1345.12_{\pm22.84}$ \\
\rowcolor[HTML]{C6F1C6} 
\textbf{TPC} & $\textbf{189.00}_{\pm3.94}$            & $\textbf{149.16}_{\pm5.68}$        & $\textbf{130.16}_{\pm8.91}$           & $\textbf{166.00}_{\pm5.16}$        & $\textbf{145.92}_{\pm2.99}$          & $\textbf{134.97}_{\pm3.52}$            & $\textbf{157.73}_{\pm1.83}$        & $\textbf{119.30}_{\pm3.39}$          & $\textbf{122.00}_{\pm12.46}$     & $\textbf{160.08}_{\pm1.40}$            & $\textbf{1474.32}_{\pm20.84}$ \\ \hline
\end{tabular}%
}
\caption{\textbf{MME Perception task}. Perception task primarily assesses the model's coarse-grained and fine-grained perceptual abilities. TPC demonstrates a strong capacity for information integration, enhancing both levels of perception.}
\label{MME Perception}
\end{table*}

\begin{table*}[htbp]
\centering
\tiny
\resizebox{\textwidth}{!}{%
\begin{tabular}{|c|cc|cccccccc|}
\hline
Methods      & \textbf{Avg$\uparrow$} & \textbf{Hall. Rate$\downarrow$} & Attribute$\uparrow$ & Adversarial$\uparrow$ & Comparison$\uparrow$ & Counting$\uparrow$ & Relation$\uparrow$ & Environment$\uparrow$ & Holistic$\uparrow$ & Other$\uparrow$ \\ \hline
Regular      & 1.72                 & 0.66                          & 2.67              & \textbf{1.75}       & 1.42               & 1.25             & 2.08             & 2.33                & 0.67             & 1.58          \\
VCD          & 1.66                 & 0.70                           & 2.33              & 1.25                & 1.42               & 0.92             & 1.83             & 1.92                & 1.00                & \textbf{2.58} \\
DoLa         & 1.92                 & 0.69                          & \textbf{2.83}     & 0.42                & 2.08               & \textbf{2.17}    & 1.42             & 2.75                & 1.33             & 2.33          \\
\rowcolor[HTML]{C6F1C6} 
\textbf{TPC} & \textbf{2.32}        & \textbf{0.62}                 & 2.17              & 1.50                 & \textbf{3.58}      & 2.08             & \textbf{2.25}    & \textbf{3.58}       & \textbf{1.58}    & 1.83          \\ \hline
\end{tabular}%
}
\caption{The results of the MMHal-Bench.}
\label{MMHal}
\end{table*}

\begin{table}[htbp]
\centering
\large
\resizebox{\columnwidth}{!}{%
\begin{tabular}{|c|cccc|c|}
\hline
Method       & \multicolumn{1}{l}{Commonsense} & \multicolumn{1}{l}{Numerical} & \multicolumn{1}{l}{Translation} & \multicolumn{1}{l|}{Code} & \textbf{Cognition score} \\ \hline
Regular      & $105.86_{\pm5.18}$              & $64.50_{\pm12.24}$            & $85.00_{\pm20.24}$              & $\textbf{79.00}_{\pm25.45}$        & $334.36_{\pm21.98}$      \\
VCD          & $107.07_{\pm14.90}$             & $\textbf{77.50}_{\pm16.07}$            & $86.00_{\pm13.60}$                 & $73.00_{\pm13.32}$        & $\textbf{342.57}_{\pm19.39}$      \\
DoLa         & $108.71_{\pm10.86}$             & $65.25_{\pm14.69}$            & $78.50_{\pm14.05}$              & $67.75_{\pm12.27}$        & $320.21_{\pm19.12}$      \\
\rowcolor[HTML]{C6F1C6} 
\textbf{TPC} & $\textbf{119.13}_{\pm5.75}$              & $50.25_{\pm7.86}$             & $\textbf{97.50}_{\pm10.27}$               & $73.00_{\pm14.62}$        & $340.68_{\pm21.73}$      \\ \hline
\end{tabular}%
}
\caption{\textbf{MME Cognition task}. Cognition is divided into four parts: commonsense reasoning, numerical calculation, text translation, and code reasoning. TPC improves reasoning abilities overall, but shows a notable decline in numerical calculation.}
\label{MME Cognition}
\end{table}

\begin{table*}[]
\centering
\resizebox{\textwidth}{!}{%
\begin{tabular}{|cccccccccccccccccccl|}
\hline
\multicolumn{19}{|c}{\textbf{POPE MSCOCO}}                                                                                                                                                                                                                                                                                                                                                                                                                                                                                                                                                                                                                                                                                                                                                  & \textbf{}                                  \\ \hline
\multicolumn{1}{|c|}{}                                     &                       & \multicolumn{4}{c}{Popular}                                                                                                                                                &                       & \multicolumn{4}{c}{Random}                                                                                                                                           &                       & \multicolumn{4}{c}{Adversarial}                                                                                                                                              & \multicolumn{1}{c|}{} &                                        &                                                    &                                            \\
\hline
\end{tabular}%
}
\caption{\textbf{Hallucination Impact.} Adding a hallucination prompt to the input significantly degrades the performance of the VLM. TPC also experiences a noticeable drop but still maintains the highest accuracy compared to the baselines, with an HI score second only to VCD.}
\label{HI}
\end{table*}

\begin{table}[htbp]
\centering
\Large
\resizebox{\columnwidth}{!}{%
\begin{tabular}{|c|cccccccccc|cc|}
\hline
\multirow{2}{*}{Window Size} &  & \multicolumn{2}{c}{Popular}    &  & \multicolumn{2}{c}{Random}      &  & \multicolumn{2}{c}{Adversarial} &  & \multirow{2}{*}{\textbf{Acc.(avg)}} & \multirow{2}{*}{\textbf{F1(avg)}} \\
                             &  & Acc.          & F1             &  & Acc.           & F1             &  & Acc.           & F1             &  &                                     &                                   \\ \hline
5                            &  & 85.73         & 84.45          &  & 87.40           & 86.01          &  & 83.20           & 82.29          &  & 85.44                               & 84.25                             \\
10                           &  & 86.27         & 85.38          &  & 88.43          & 87.41          &  & 83.00             & 82.58          &  & 85.90                               & 85.12                             \\
15                           &  & \textbf{86.40} & \textbf{85.55} &  & 88.50           & 87.51          &  & \textbf{83.10}  & \textbf{82.73} &  & \textbf{86.00}                      & 85.26                             \\
\textbf{20}                  &  & \textbf{86.40} & \textbf{85.55} &  & \textbf{88.53} & \textbf{87.56} &  & 83.07          & 82.70           &  & \textbf{86.00}                      & \textbf{85.27}                    \\ \hline
\end{tabular}%
}
\caption{\textbf{Ablate window size.} Connecting the logits at the end of the input sequence and gradually increasing the window size.}
\label{tab:window size}
\end{table}

\paragraph{Benchmark \& Evaluation Metrics}
CHAIR \cite{rohrbach2018object}, a popular metric for image captioning, works by generating a ground-truth object label set for each image. Objects in the caption but absent from the label set are deemed hallucinated. CHAIR has two evaluation dimensions: instance-level (CHAIRi) and sentence-level (CHAIRs).

\begin{equation}
    \text{CHAIRi}=\frac{|\text{\{hallucinated objects\}}|}{\text{all mentioned objects}}
\end{equation}

\begin{equation}
    \text{CHAIRs}=\frac{|\text{\{captions with hallucinated objects\}}|}{\text{all captions}}
\end{equation}

\paragraph{Models \& Baselines.}
We select three recent, highly effective representatives in CD: 1) VCD~\cite{leng2024mitigating}, a prominent example of CD in the multimodal domain, 2) DoLa~\cite{chuang2023dola}, a powerful recent baseline that requires no training and offers nearly lossless inference speed, and 3) PAI \cite{liu2024paying}, a method for enhancing the model's attention to image. It should be noted that DoLa was originally designed to mitigate hallucinations in single-modal LLMs, but we observe that it also performs well in reducing multimodal hallucinations. DoLa supports only LLaMA-based models and is therefore excluded from QwenVL comparisons. The results presented are based on our multimodal implementation, where we select all odd-numbered layers as candidate premature layers, keeping other hyperparameters consistent with the original paper. Our base models are LLaVA1.5-7b, LLaVA1.5-13b \cite{liu2023llava}, QwenVL \cite{Qwen-VL} and MiniGPT4 \cite{zhu2023minigpt}, and the baselines include the nucleus sampling methods, as well as VCD and DoLa. Unless otherwise specified, the hyperparameter settings used in this paper are \( \lambda = 0.1 \), \( \alpha = 3.0 \)\footnote{In practical experiments, before connecting the current logits temporally, we multiply these logits by \(\alpha\) to prevent excessive influence from the logits of preceding timesteps, followed by normalization.}, and Temperature = 1.0.

\subsection{Main results}
\paragraph{Object Hallucination.} We assess how effectively TPC and other baselines mitigate object hallucination on the POPE benchmark. As shown in Table \ref{POPE All}, we evaluate direct sampling (hereafter referred to as Regular, the comparison with greedy search and nucleus sampling with different hyper-parameters is presented in Supplementary \ref{sec: greedy nucleus}), VCD, DoLa, and our TPC’s performance on POPE with LLaVA1.5-7\&13b and QwenVL. On all three POPE splits—Popular, Random, and Adversarial—TPC achieves higher accuracy and F1 scores than the baselines. Taking LLaVA1.5-7b as an example, compared to Regular, TPC demonstrates an average accuracy improvement of 3.52\%-5.19\% and an F1 score increase of 3.57\%–5.12\%. When compared to the strongest baseline, DoLa, TPC increases average accuracy by 0.48\%–1.70\% and average F1 score by 1.01\%–1.68\%. Switching to a larger model like LLaVA1.5-13b or a more powerful VLM QwenVL, TPC remains effective, supporting the robustness of the approach. In Table \ref{minigpt}, 500 images are randomly selected from MSCOCO, and MiniGPT4 is employed for the image captioning task. The CHAIR metric is then compared with that of PAI. The results of PAI are obtained by implementing the parameters set in the original paper. Although the CHAIR score of TPC is higher than that of PAI, it is found that combining the two can lead to better results. We also compare the Accuracy on POPE between TPC and PAI. Under the two decoding strategies of greedy search and beam search, TPC has a slight advantage. 

Table \ref{MME Perception} shows the results on MME, where the Existence and Count splits focus on object-level hallucination detection, while the Position and Color splits pertain to attribute-level. TPC significantly improves the VLM’s scores across both levels. These results suggest that rather than removing harmful logits information through contrastive decoding, leveraging logits from different time steps enriches the contextual information for the next token, effectively reducing the model’s tendency toward errors. TPC demonstrates a promising alternative for mitigating object hallucination.
\\

\noindent \textbf{Perception \& Cognition.}
Perception and Cognition tasks respectively measure the VLMs' ability to perceive images and to reason based on images. We evaluate TPC's performance across the entire MME dataset. Due to considerable variability in MME test data, we conduct 10 consecutive tests and averaged the results. In the Perception task, as shown in Table \ref{MME Perception}, it is evident that TPC outperforms other baselines across all Perception splits. This improvement in perceptual is likely attributable to TPC, where each generated token directly references all input tokens, enhancing the model’s ability to perceive contextual information.

In the Cognition tasks, TPC exhibits significant enhancements in commonsense reasoning and translation. As presented in Table \ref{MME Cognition}, while TPC yields overall improvements in cognition compared to the Regular baseline, a performance decrement is observed in code-related capabilities. This pattern is consistent with findings observed in VCD and DoLa. The observed decline may stem from the fact that, although TPC augments contextual information for each generated token, tasks requiring intricate reasoning and high logical demands can be negatively affected. These logical capabilities, primarily acquired during training, prove challenging to further refine through post-hoc methodologies.\\

\noindent \textbf{Open-Ended Text Generation.} We observe that while certain decoding methods perform well on the datasets mentioned above, they may degrade the quality of the generated text, sometimes even increasing the proportion of hallucinated tokens. Using MMHal-Bench, we generate responses and have GPT-4 \cite{achiam2023gpt} evaluate these against the ground truth for information richness and hallucination rate. As shown in Table \ref{MMHal}, VCD results in a slight decrease in information richness and a marginal increase in hallucination rate. DoLa improves information richness significantly but also leads to a notable increase in hallucination. TPC, however, not only enhances information further but also effectively curbs the hallucination rate, demonstrating its suitability for open-ended text generation.

\begin{figure*}[htbp]
    \centering
    \includegraphics[width=\linewidth]{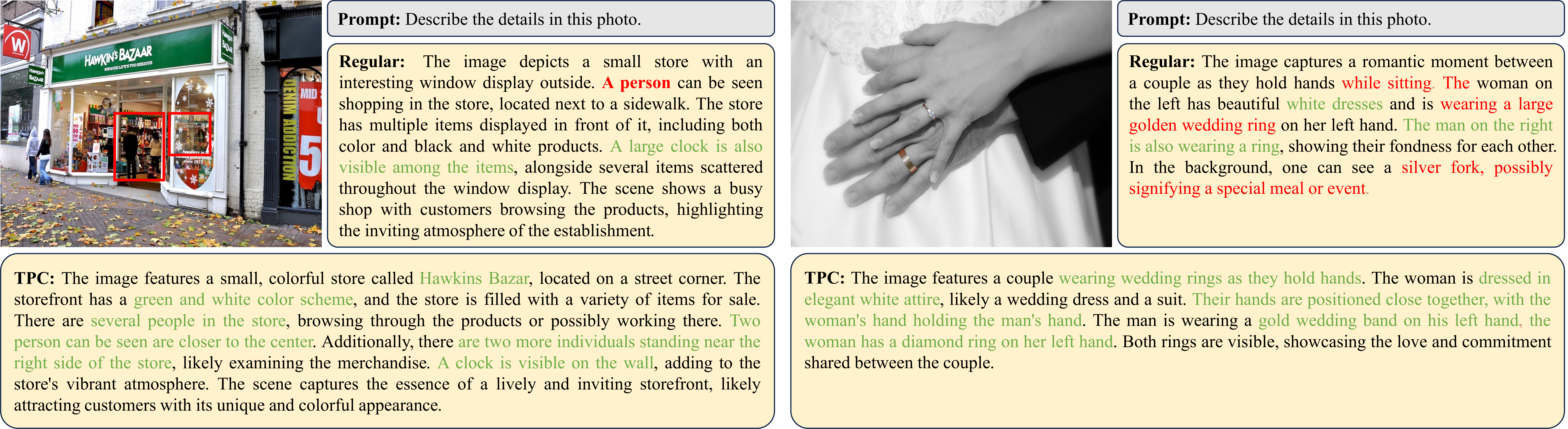}
    \caption{\textbf{Qualitative analysis.} The cases are taken from MMHal-Bench, with \textcolor{green}{green text} indicating valuable information. In the left image case, the output from LLaVA1.5-7b is shown, which, despite containing only a small amount of hallucination, lacks sufficiently detailed information. In contrast, TPC demonstrates fine-grained perceptual ability, even identifying two individuals behind the store window (in the right-side bounding box).}
    \label{fig:cases}
\end{figure*}

\subsection{Hallucination Impact}
According to studies by \cite{zhong2024investigating,favero2024multi}, models are highly susceptible to generating hallucinated tokens when the input contains hallucination. To investigate this, we conduct a simple experiment by appending a hallucination-inducing prompt, “\textit{There is nothing in the image}” to the end of the input. This prompt naturally degrades model performance, and we initially hypothesize that TPC’s performance might deteriorate more severely than the baselines since the output tokens are adjacent to the hallucination prompt and thus more prone to its influence. We design a simple metric, termed Hallucination Impact (HI), as follows:
\begin{equation}
    \text{HI} = Acc_{origin} - Acc_{hallu},     
\end{equation}
where $Acc_{origin}$ and $Acc_{hallu}$ represent the accuracy without hallucinated token input and with hallucinated token input, respectively. Testing on POPE MSCOCO, however, reveals an interesting result in Table \ref{HI}: TPC still outperforms the baselines in performance, with an HI score of 11.02, second only to VCD's 10.22. It likely stems from the fact that, although hallucination affects the model, the connection of logits incorporates authentic knowledge from preceding tokens. While the attenuation factor $\lambda$ causes this knowledge to weaken over distance, the impact of hallucination remains localized. We believe that VCD achieves the lowest HI score because it inherently performs contrastive decoding with hallucinated logits, making it less sensitive to hallucinated tokens.

\subsection{Ablation}
In Table \ref{tab:window size}, we present an ablation study with different window sizes. We consistently set the last input token as the focal point of the window and adjust the window size to ensure that the logits within the window form a continuous sequence leading up to the output token. Results show that as the window size increases, the accuracy also gradually improves, as more information from the sequence is connected to the output token. However, due to $\lambda$, the logits earlier in the sequence are attenuated, causing accuracy to gradually converge.

\subsection{Qualitative Study}
We select text descriptions generated by Regular and TPC from MMHal-Bench for analysis. As shown in Figure \ref{fig:cases}, green indicates factual information, while red denotes hallucinations. In the left image, Regular introduces a hallucination, observing only one person in the store. In contrast, TPC not only accurately identifies the location of the customer in the store but also detects two individuals behind the display window on the right side of the image, as indicated by the red bounding box. In the right image, despite the limited visual information, Regular generates numerous hallucinated tokens, whereas TPC not only avoids any hallucinated descriptions but is also highly informative. Additional cases can be found in Supplementary \ref{sec: cases}.
\section{Conclusion}
Building on our observation of the logits' continuity consistency enhancement characteristic, we observe the divergence variation of logits at different timestep distances, and how connecting adjacent logits improves their semantic consistency through context incorporation across time steps. Motivated by this insight, we propose a straightforward, training-free, and plug-and-play approach called Cross-Temporal Prediction Connection (TPC), which leverages temporal connections to integrate logits across adjacent time steps. By connecting logits over time, TPC not only strengthens the coherence and robustness of the model’s output but also enriches its contextual understanding. Our experimental results confirm that TPC effectively reduces hallucinations, improving the informativeness, and contextual relevance of outputs.\\

\noindent \textbf{Future Work.} In this study, we focus exclusively on connecting logits and do not extend our analysis to temporal dependencies at the hidden state level. As part of our future work, we plan to conduct a more comprehensive investigation into the effects of establishing connections between hidden states across different timesteps. \\
{
    \small
    \bibliographystyle{ieeenat_fullname}
    \bibliography{main}

\begin{thebibliography}{49}
\providecommand{\natexlab}[1]{#1}
\providecommand{\url}[1]{\texttt{#1}}
\expandafter\ifx\csname urlstyle\endcsname\relax
  \providecommand{\doi}[1]{doi: #1}\else
  \providecommand{\doi}{doi: \begingroup \urlstyle{rm}\Url}\fi

\bibitem[Abdi and Williams(2010)]{abdi2010principal}
Herv{\'e} Abdi and Lynne~J Williams.
\newblock Principal component analysis.
\newblock \emph{Wiley interdisciplinary reviews: computational statistics}, 2\penalty0 (4):\penalty0 433--459, 2010.

\bibitem[Achiam et~al.(2023)Achiam, Adler, Agarwal, Ahmad, Akkaya, Aleman, Almeida, Altenschmidt, Altman, Anadkat, et~al.]{achiam2023gpt}
Josh Achiam, Steven Adler, Sandhini Agarwal, Lama Ahmad, Ilge Akkaya, Florencia~Leoni Aleman, Diogo Almeida, Janko Altenschmidt, Sam Altman, Shyamal Anadkat, et~al.
\newblock Gpt-4 technical report.
\newblock \emph{arXiv preprint arXiv:2303.08774}, 2023.

\bibitem[Bai et~al.(2023)Bai, Bai, Yang, Wang, Tan, Wang, Lin, Zhou, and Zhou]{Qwen-VL}
Jinze Bai, Shuai Bai, Shusheng Yang, Shijie Wang, Sinan Tan, Peng Wang, Junyang Lin, Chang Zhou, and Jingren Zhou.
\newblock Qwen-vl: A frontier large vision-language model with versatile abilities.
\newblock \emph{arXiv preprint arXiv:2308.12966}, 2023.

\bibitem[Bai et~al.(2024)Bai, Wang, Xiao, He, Han, Zhang, and Shou]{bai2024hallucination}
Zechen Bai, Pichao Wang, Tianjun Xiao, Tong He, Zongbo Han, Zheng Zhang, and Mike~Zheng Shou.
\newblock Hallucination of multimodal large language models: A survey.
\newblock \emph{arXiv preprint arXiv:2404.18930}, 2024.

\bibitem[Bhadra et~al.(2021)Bhadra, Kelkar, Brooks, and Anastasio]{bhadra2021hallucinations}
Sayantan Bhadra, Varun~A Kelkar, Frank~J Brooks, and Mark~A Anastasio.
\newblock On hallucinations in tomographic image reconstruction.
\newblock \emph{IEEE transactions on medical imaging}, 40\penalty0 (11):\penalty0 3249--3260, 2021.

\bibitem[Chen et~al.(2024{\natexlab{a}})Chen, Xiong, Liu, Wu, Xiao, Gao, and He]{chen2024context}
Shiqi Chen, Miao Xiong, Junteng Liu, Zhengxuan Wu, Teng Xiao, Siyang Gao, and Junxian He.
\newblock In-context sharpness as alerts: An inner representation perspective for hallucination mitigation.
\newblock \emph{arXiv preprint arXiv:2403.01548}, 2024{\natexlab{a}}.

\bibitem[Chen et~al.(2024{\natexlab{b}})Chen, Wu, Wang, Su, Chen, Xing, Zhong, Zhang, Zhu, Lu, et~al.]{chen2024internvl}
Zhe Chen, Jiannan Wu, Wenhai Wang, Weijie Su, Guo Chen, Sen Xing, Muyan Zhong, Qinglong Zhang, Xizhou Zhu, Lewei Lu, et~al.
\newblock Internvl: Scaling up vision foundation models and aligning for generic visual-linguistic tasks.
\newblock In \emph{Proceedings of the IEEE/CVF Conference on Computer Vision and Pattern Recognition}, pages 24185--24198, 2024{\natexlab{b}}.

\bibitem[Chuang et~al.(2023)Chuang, Xie, Luo, Kim, Glass, and He]{chuang2023dola}
Yung-Sung Chuang, Yujia Xie, Hongyin Luo, Yoon Kim, James Glass, and Pengcheng He.
\newblock Dola: Decoding by contrasting layers improves factuality in large language models.
\newblock \emph{arXiv preprint arXiv:2309.03883}, 2023.

\bibitem[Cohen et~al.(2018)Cohen, Luck, and Honari]{cohen2018distribution}
Joseph~Paul Cohen, Margaux Luck, and Sina Honari.
\newblock Distribution matching losses can hallucinate features in medical image translation.
\newblock In \emph{Medical Image Computing and Computer Assisted Intervention--MICCAI 2018: 21st International Conference, Granada, Spain, September 16-20, 2018, Proceedings, Part I}, pages 529--536. Springer, 2018.

\bibitem[Dai et~al.(2024)Dai, Guo, Tang, Li, Wang, Huang, Tian, Xia, Lv, and Wang]{dai2024vistarag}
Xingyuan Dai, Chao Guo, Yun Tang, Haichuan Li, Yutong Wang, Jun Huang, Yonglin Tian, Xin Xia, Yisheng Lv, and Fei-Yue Wang.
\newblock Vistarag: Toward safe and trustworthy autonomous driving through retrieval-augmented generation.
\newblock \emph{IEEE Transactions on Intelligent Vehicles}, 2024.

\bibitem[Del{\'e}tang et~al.(2023)Del{\'e}tang, Ruoss, Duquenne, Catt, Genewein, Mattern, Grau-Moya, Wenliang, Aitchison, Orseau, et~al.]{deletang2023language}
Gr{\'e}goire Del{\'e}tang, Anian Ruoss, Paul-Ambroise Duquenne, Elliot Catt, Tim Genewein, Christopher Mattern, Jordi Grau-Moya, Li~Kevin Wenliang, Matthew Aitchison, Laurent Orseau, et~al.
\newblock Language modeling is compression.
\newblock \emph{arXiv preprint arXiv:2309.10668}, 2023.

\bibitem[Diao et~al.(2024)Diao, Cui, Li, Wang, Lu, and Wang]{diao2024EVE}
Haiwen Diao, Yufeng Cui, Xiaotong Li, Yueze Wang, Huchuan Lu, and Xinlong Wang.
\newblock Unveiling encoder-free vision-language models.
\newblock \emph{arXiv preprint arXiv:2406.11832}, 2024.

\bibitem[Favero et~al.(2024)Favero, Zancato, Trager, Choudhary, Perera, Achille, Swaminathan, and Soatto]{favero2024multi}
Alessandro Favero, Luca Zancato, Matthew Trager, Siddharth Choudhary, Pramuditha Perera, Alessandro Achille, Ashwin Swaminathan, and Stefano Soatto.
\newblock Multi-modal hallucination control by visual information grounding.
\newblock In \emph{Proceedings of the IEEE/CVF Conference on Computer Vision and Pattern Recognition}, pages 14303--14312, 2024.

\bibitem[Fu et~al.(2023)Fu, Chen, Shen, Qin, Zhang, Lin, Yang, Zheng, Li, Sun, et~al.]{fu2023mme}
Chaoyou Fu, Peixian Chen, Yunhang Shen, Yulei Qin, Mengdan Zhang, Xu Lin, Jinrui Yang, Xiawu Zheng, Ke Li, Xing Sun, et~al.
\newblock Mme: A comprehensive evaluation benchmark for multimodal large language models.
\newblock \emph{arXiv preprint arXiv:2306.13394}, 2023.

\bibitem[Guan et~al.(2024)Guan, Liu, Wu, Xian, Li, Liu, Wang, Chen, Huang, Yacoob, et~al.]{guan2024hallusionbench}
Tianrui Guan, Fuxiao Liu, Xiyang Wu, Ruiqi Xian, Zongxia Li, Xiaoyu Liu, Xijun Wang, Lichang Chen, Furong Huang, Yaser Yacoob, et~al.
\newblock Hallusionbench: an advanced diagnostic suite for entangled language hallucination and visual illusion in large vision-language models.
\newblock In \emph{Proceedings of the IEEE/CVF Conference on Computer Vision and Pattern Recognition}, pages 14375--14385, 2024.

\bibitem[Gunjal et~al.(2024)Gunjal, Yin, and Bas]{gunjal2024detecting}
Anisha Gunjal, Jihan Yin, and Erhan Bas.
\newblock Detecting and preventing hallucinations in large vision language models.
\newblock In \emph{Proceedings of the AAAI Conference on Artificial Intelligence}, pages 18135--18143, 2024.

\bibitem[Gupta et~al.(2022)Gupta, Li, Kortylewski, Zhang, Li, and Yuille]{gupta2022swapmix}
Vipul Gupta, Zhuowan Li, Adam Kortylewski, Chenyu Zhang, Yingwei Li, and Alan Yuille.
\newblock Swapmix: Diagnosing and regularizing the over-reliance on visual context in visual question answering.
\newblock In \emph{Proceedings of the IEEE/CVF conference on computer vision and pattern recognition}, pages 5078--5088, 2022.

\bibitem[Hamidieh et~al.(2024)Hamidieh, Zhang, Gerych, Hartvigsen, and Ghassemi]{hamidieh2024identifying}
Kimia Hamidieh, Haoran Zhang, Walter Gerych, Thomas Hartvigsen, and Marzyeh Ghassemi.
\newblock Identifying implicit social biases in vision-language models.
\newblock In \emph{Proceedings of the AAAI/ACM Conference on AI, Ethics, and Society}, pages 547--561, 2024.

\bibitem[Hu et~al.(2023)Hu, Zhang, Zhao, and Sun]{hu2023ciem}
Hongyu Hu, Jiyuan Zhang, Minyi Zhao, and Zhenbang Sun.
\newblock Ciem: Contrastive instruction evaluation method for better instruction tuning.
\newblock \emph{arXiv preprint arXiv:2309.02301}, 2023.

\bibitem[Huang et~al.(2024)Huang, Dong, Zhang, Wang, He, Wang, Lin, Zhang, and Yu]{huang2024opera}
Qidong Huang, Xiaoyi Dong, Pan Zhang, Bin Wang, Conghui He, Jiaqi Wang, Dahua Lin, Weiming Zhang, and Nenghai Yu.
\newblock Opera: Alleviating hallucination in multi-modal large language models via over-trust penalty and retrospection-allocation.
\newblock In \emph{Proceedings of the IEEE/CVF Conference on Computer Vision and Pattern Recognition}, pages 13418--13427, 2024.

\bibitem[Jiang et~al.(2024)Jiang, Xu, Dong, Chen, Ye, Yan, Ye, Zhang, Huang, and Zhang]{jiang2024hallucination}
Chaoya Jiang, Haiyang Xu, Mengfan Dong, Jiaxing Chen, Wei Ye, Ming Yan, Qinghao Ye, Ji Zhang, Fei Huang, and Shikun Zhang.
\newblock Hallucination augmented contrastive learning for multimodal large language model.
\newblock In \emph{Proceedings of the IEEE/CVF Conference on Computer Vision and Pattern Recognition}, pages 27036--27046, 2024.

\bibitem[Leng et~al.(2024)Leng, Zhang, Chen, Li, Lu, Miao, and Bing]{leng2024mitigating}
Sicong Leng, Hang Zhang, Guanzheng Chen, Xin Li, Shijian Lu, Chunyan Miao, and Lidong Bing.
\newblock Mitigating object hallucinations in large vision-language models through visual contrastive decoding.
\newblock In \emph{Proceedings of the IEEE/CVF Conference on Computer Vision and Pattern Recognition}, pages 13872--13882, 2024.

\bibitem[Li et~al.(2020)Li, Yin, Li, Hu, Zhang, Zhang, Wang, Hu, Dong, Wei, Choi, and Gao]{li2020oscar}
Xiujun Li, Xi Yin, Chunyuan Li, Xiaowei Hu, Pengchuan Zhang, Lei Zhang, Lijuan Wang, Houdong Hu, Li Dong, Furu Wei, Yejin Choi, and Jianfeng Gao.
\newblock Oscar: Object-semantics aligned pre-training for vision-language tasks.
\newblock \emph{ECCV 2020}, 2020.

\bibitem[Li et~al.(2022)Li, Holtzman, Fried, Liang, Eisner, Hashimoto, Zettlemoyer, and Lewis]{li2022contrastive}
Xiang~Lisa Li, Ari Holtzman, Daniel Fried, Percy Liang, Jason Eisner, Tatsunori Hashimoto, Luke Zettlemoyer, and Mike Lewis.
\newblock Contrastive decoding: Open-ended text generation as optimization.
\newblock \emph{arXiv preprint arXiv:2210.15097}, 2022.

\bibitem[Li et~al.(2023)Li, Du, Zhou, Wang, Zhao, and Wen]{li2023evaluating}
Yifan Li, Yifan Du, Kun Zhou, Jinpeng Wang, Wayne~Xin Zhao, and Ji-Rong Wen.
\newblock Evaluating object hallucination in large vision-language models.
\newblock \emph{arXiv preprint arXiv:2305.10355}, 2023.

\bibitem[Liu et~al.(2023{\natexlab{a}})Liu, Lin, Li, Wang, Yacoob, and Wang]{liu2023aligning}
Fuxiao Liu, Kevin Lin, Linjie Li, Jianfeng Wang, Yaser Yacoob, and Lijuan Wang.
\newblock Aligning large multi-modal model with robust instruction tuning.
\newblock \emph{arXiv preprint arXiv:2306.14565}, 2023{\natexlab{a}}.

\bibitem[Liu et~al.(2023{\natexlab{b}})Liu, Lin, Li, Wang, Yacoob, and Wang]{liu2023mitigating}
Fuxiao Liu, Kevin Lin, Linjie Li, Jianfeng Wang, Yaser Yacoob, and Lijuan Wang.
\newblock Mitigating hallucination in large multi-modal models via robust instruction tuning.
\newblock In \emph{The Twelfth International Conference on Learning Representations}, 2023{\natexlab{b}}.

\bibitem[Liu et~al.(2023{\natexlab{c}})Liu, Li, Wu, and Lee]{liu2023llava}
Haotian Liu, Chunyuan Li, Qingyang Wu, and Yong~Jae Lee.
\newblock Visual instruction tuning, 2023{\natexlab{c}}.

\bibitem[Liu et~al.(2024)Liu, Zheng, and Chen]{liu2024paying}
Shi Liu, Kecheng Zheng, and Wei Chen.
\newblock Paying more attention to image: A training-free method for alleviating hallucination in lvlms.
\newblock \emph{arXiv preprint arXiv:2407.21771}, 2024.

\bibitem[Raffel et~al.(2020)Raffel, Shazeer, Roberts, Lee, Narang, Matena, Zhou, Li, and Liu]{raffel2020exploring}
Colin Raffel, Noam Shazeer, Adam Roberts, Katherine Lee, Sharan Narang, Michael Matena, Yanqi Zhou, Wei Li, and Peter~J Liu.
\newblock Exploring the limits of transfer learning with a unified text-to-text transformer.
\newblock \emph{Journal of machine learning research}, 21\penalty0 (140):\penalty0 1--67, 2020.

\bibitem[Rawte et~al.(2023)Rawte, Sheth, and Das]{rawte2023survey}
Vipula Rawte, Amit Sheth, and Amitava Das.
\newblock A survey of hallucination in large foundation models.
\newblock \emph{arXiv preprint arXiv:2309.05922}, 2023.

\bibitem[Rohrbach et~al.(2018)Rohrbach, Hendricks, Burns, Darrell, and Saenko]{rohrbach2018object}
Anna Rohrbach, Lisa~Anne Hendricks, Kaylee Burns, Trevor Darrell, and Kate Saenko.
\newblock Object hallucination in image captioning.
\newblock \emph{arXiv preprint arXiv:1809.02156}, 2018.

\bibitem[Sun et~al.(2023)Sun, Shen, Cao, Liu, Li, Shen, Gan, Gui, Wang, Yang, et~al.]{sun2023aligning}
Zhiqing Sun, Sheng Shen, Shengcao Cao, Haotian Liu, Chunyuan Li, Yikang Shen, Chuang Gan, Liang-Yan Gui, Yu-Xiong Wang, Yiming Yang, et~al.
\newblock Aligning large multimodal models with factually augmented rlhf.
\newblock \emph{arXiv preprint arXiv:2309.14525}, 2023.

\bibitem[Touvron et~al.(2023)Touvron, Martin, Stone, Albert, Almahairi, Babaei, Bashlykov, Batra, Bhargava, Bhosale, et~al.]{touvron2023llama}
Hugo Touvron, Louis Martin, Kevin Stone, Peter Albert, Amjad Almahairi, Yasmine Babaei, Nikolay Bashlykov, Soumya Batra, Prajjwal Bhargava, Shruti Bhosale, et~al.
\newblock Llama 2: Open foundation and fine-tuned chat models.
\newblock \emph{arXiv preprint arXiv:2307.09288}, 2023.

\bibitem[Wang et~al.(2023)Wang, Wang, Xu, Zhang, Gu, Jia, Yan, Zhang, and Sang]{wang2023llm}
Junyang Wang, Yuhang Wang, Guohai Xu, Jing Zhang, Yukai Gu, Haitao Jia, Ming Yan, Ji Zhang, and Jitao Sang.
\newblock An llm-free multi-dimensional benchmark for mllms hallucination evaluation.
\newblock \emph{arXiv preprint arXiv:2311.07397}, 2023.

\bibitem[Wang et~al.(2024)Wang, Pan, Ding, and Biemann]{wang2024mitigating}
Xintong Wang, Jingheng Pan, Liang Ding, and Chris Biemann.
\newblock Mitigating hallucinations in large vision-language models with instruction contrastive decoding.
\newblock \emph{arXiv preprint arXiv:2403.18715}, 2024.

\bibitem[Wu et~al.(2023)Wu, Li, Ermon, Haffner, Ge, and Zhang]{wu2023role}
Chenwei Wu, Li~Erran Li, Stefano Ermon, Patrick Haffner, Rong Ge, and Zaiwei Zhang.
\newblock The role of linguistic priors in measuring compositional generalization of vision-language models.
\newblock In \emph{Proceedings on}, pages 118--126. PMLR, 2023.

\bibitem[Wu et~al.(2024)Wu, Liu, Wang, Zhang, Wu, Wang, and Tan]{wu2024logical}
Junfei Wu, Qiang Liu, Ding Wang, Jinghao Zhang, Shu Wu, Liang Wang, and Tieniu Tan.
\newblock Logical closed loop: Uncovering object hallucinations in large vision-language models.
\newblock \emph{arXiv preprint arXiv:2402.11622}, 2024.

\bibitem[Yin et~al.(2023)Yin, Fu, Zhao, Xu, Wang, Sui, Shen, Li, Sun, and Chen]{yin2023woodpecker}
Shukang Yin, Chaoyou Fu, Sirui Zhao, Tong Xu, Hao Wang, Dianbo Sui, Yunhang Shen, Ke Li, Xing Sun, and Enhong Chen.
\newblock Woodpecker: Hallucination correction for multimodal large language models.
\newblock \emph{arXiv preprint arXiv:2310.16045}, 2023.

\bibitem[Yu et~al.(2024)Yu, Li, Wei, Pang, Ye, Qin, Tang, Tian, and Zhuang]{yu2024hallucidoctor}
Qifan Yu, Juncheng Li, Longhui Wei, Liang Pang, Wentao Ye, Bosheng Qin, Siliang Tang, Qi Tian, and Yueting Zhuang.
\newblock Hallucidoctor: Mitigating hallucinatory toxicity in visual instruction data.
\newblock In \emph{Proceedings of the IEEE/CVF Conference on Computer Vision and Pattern Recognition}, pages 12944--12953, 2024.

\bibitem[Yue et~al.(2024)Yue, Zhang, and Jin]{yue2024less}
Zihao Yue, Liang Zhang, and Qin Jin.
\newblock Less is more: Mitigating multimodal hallucination from an eos decision perspective.
\newblock \emph{arXiv preprint arXiv:2402.14545}, 2024.

\bibitem[Zhang et~al.(2024{\natexlab{a}})Zhang, Huang, Jin, and Lu]{10445007}
Jingyi Zhang, Jiaxing Huang, Sheng Jin, and Shijian Lu.
\newblock Vision-language models for vision tasks: A survey.
\newblock \emph{IEEE Transactions on Pattern Analysis and Machine Intelligence}, 46\penalty0 (8):\penalty0 5625--5644, 2024{\natexlab{a}}.

\bibitem[Zhang et~al.(2024{\natexlab{b}})Zhang, Huang, Jin, and Lu]{zhang2024vision}
Jingyi Zhang, Jiaxing Huang, Sheng Jin, and Shijian Lu.
\newblock Vision-language models for vision tasks: A survey.
\newblock \emph{IEEE Transactions on Pattern Analysis and Machine Intelligence}, 2024{\natexlab{b}}.

\bibitem[Zhang et~al.(2022)Zhang, Wang, and Sang]{zhang2022counterfactually}
Yi Zhang, Junyang Wang, and Jitao Sang.
\newblock Counterfactually measuring and eliminating social bias in vision-language pre-training models.
\newblock In \emph{Proceedings of the 30th ACM International Conference on Multimedia}, pages 4996--5004, 2022.

\bibitem[Zhang et~al.(2024{\natexlab{c}})Zhang, Yu, Wen, Wang, Zhang, Wang, Jin, and Tan]{zhang2024debiasing}
Yi-Fan Zhang, Weichen Yu, Qingsong Wen, Xue Wang, Zhang Zhang, Liang Wang, Rong Jin, and Tieniu Tan.
\newblock Debiasing large visual language models.
\newblock \emph{arXiv preprint arXiv:2403.05262}, 2024{\natexlab{c}}.

\bibitem[Zhong et~al.(2024)Zhong, Feng, Zhao, Li, Huang, Gu, Ma, Xu, and Qin]{zhong2024investigating}
Weihong Zhong, Xiaocheng Feng, Liang Zhao, Qiming Li, Lei Huang, Yuxuan Gu, Weitao Ma, Yuan Xu, and Bing Qin.
\newblock Investigating and mitigating the multimodal hallucination snowballing in large vision-language models.
\newblock \emph{arXiv preprint arXiv:2407.00569}, 2024.

\bibitem[Zhou et~al.(2022)Zhou, Yang, Loy, and Liu]{zhou2022learning}
Kaiyang Zhou, Jingkang Yang, Chen~Change Loy, and Ziwei Liu.
\newblock Learning to prompt for vision-language models.
\newblock \emph{International Journal of Computer Vision}, 130\penalty0 (9):\penalty0 2337--2348, 2022.

\bibitem[Zhou et~al.(2023)Zhou, Cui, Yoon, Zhang, Deng, Finn, Bansal, and Yao]{zhou2023analyzing}
Yiyang Zhou, Chenhang Cui, Jaehong Yoon, Linjun Zhang, Zhun Deng, Chelsea Finn, Mohit Bansal, and Huaxiu Yao.
\newblock Analyzing and mitigating object hallucination in large vision-language models.
\newblock \emph{arXiv preprint arXiv:2310.00754}, 2023.

\bibitem[Zhu et~al.(2023)Zhu, Chen, Shen, Li, and Elhoseiny]{zhu2023minigpt}
Deyao Zhu, Jun Chen, Xiaoqian Shen, Xiang Li, and Mohamed Elhoseiny.
\newblock Minigpt-4: Enhancing vision-language understanding with advanced large language models.
\newblock \emph{arXiv preprint arXiv:2304.10592}, 2023.

\end{thebibliography}
}

\clearpage
\setcounter{page}{1}
\maketitlesupplementary
\appendix

\noindent Section \ref{sec:hyperpara} of the appendix presents tuning experiments for the two hyperparameters \(\alpha\) and \(\lambda\) introduced in TPC. Section \ref{sec: efficiency} evaluates the efficiency of baselines and TPC in terms of throughput, latency, and GPU memory usage. Additionally, in section \ref{sec: greedy nucleus}, we conduct extensive experiments to compare performance when combined with greedy search and nucleus sampling. Finally, additional cases obtained from MMHal-Bench demonstrate the effectiveness of our method in section \ref{sec: cases}. The experiments are conducted with V100 GPUs.

\section{Impact of $\alpha$ and Attenuation Coefficient $\lambda$}
\label{sec:hyperpara}
To investigate the effects of $\alpha$ and $\lambda$, we first conduct a parameter-tuning experiment on POPE MSCOCO, as shown in Table \ref{tab: hyperparam}. We initially set $\alpha$ to 1.0 and gradually increases $\lambda$ from 0.1 up to 0.8. We observe that performance continuously improves with increasing $\lambda$, and the rate of improvement accelerates. Accuracy reaches its peak at $\lambda = 0.7$, only beginning to decline at 0.8. We believe this improvement is due to increased $\lambda$, which greatly enhances the consistency and contextual information in the logits connection. However, we also find that an excessively high $\lambda$ leads to a large amount of repetition in long text generation, limiting the model’s capacity to handle more complex tasks. To ensure robustness, we set $\lambda$ to 0.1 in most experiments.

We also experiment with setting $\alpha$ to 3.0, which functions similarly to the temperature in direct sampling, except that here, $\alpha$ is applied before connecting and then followed by normalization. As shown in Table \ref{tab: hyperparam}, applying $\alpha$ further improves model performance, and experiments reveal that it effectively reduces repetition. This is likely because, even when connected with other logits, a higher $\alpha$ ensures that the current time step's logits remain dominant, preserving the information without excessive distortion.

\begin{table}[]
\centering
\resizebox{\columnwidth}{!}{%
\begin{tabular}{|cc|cccccc|cc|}
\hline
\multirow{2}{*}{$\alpha$} & \multirow{2}{*}{$\lambda$} & \multicolumn{2}{c}{Popular}     & \multicolumn{2}{c}{Random}      & \multicolumn{2}{c|}{Adversarial} & \multirow{2}{*}{Acc.(avg)} & \multirow{2}{*}{F1(avg)} \\ \cline{3-8}
                          &                            & Acc.           & F1             & Acc.           & F1             & Acc.            & F1             &                            &                          \\ \hline
\multirow{9}{*}{1.0}      & 0.10                        & 83.43          & 81.54          & 84.63          & 82.65          & 80.53           & 78.96          & 82.86                      & 81.05                    \\
                          & 0.20                        & 83.77          & 81.90           & 84.93          & 82.98          & 80.90            & 79.34          & 83.20                       & 81.41                    \\
                          & 0.30                        & 84.00             & 82.18          & 85.37          & 83.45          & 81.33           & 79.78          & 83.57                      & 81.80                     \\
                          & 0.40                        & 84.77          & 83.09          & 86.03          & 84.28          & 82.27           & 80.85          & 84.36                      & 82.74                    \\
                          & 0.50                        & 84.90           & 83.34          & 86.30           & 84.65          & 82.10            & 80.84          & 84.43                      & 82.94                    \\
                          & 0.60                        & 85.73          & 84.53          & 87.47          & 86.15          & 82.47  & 81.60           & 85.22                      & 84.09                    \\
                          & 0.70                        & 86.40           & 85.52          & 88.63          & 87.68          & \textbf{83.17}           & 82.81          & 86.07                      & 85.34                    \\
                          & 0.80                        & 86.97          & 86.66          & 88.83          & 88.22          & 82.00              & 82.39          & 85.93                      & 85.76                    \\
                          & 0.90                        & \textbf{87.07} & \textbf{87.05} & \textbf{89.67} & \textbf{89.34} & 81.57           & \textbf{82.51} & \textbf{86.10}              & \textbf{86.30}            \\ \hline
\multirow{9}{*}{3.0}      & 0.10                        & 85.40           & 83.83          & 86.63          & 84.99          & 82.93           & 81.57          & 84.99                      & 83.46                    \\
                          & 0.20                        & 85.33          & 83.73          & 86.50           & 84.83          & 82.97           & 81.57          & 84.93                      & 83.38                    \\
                          & 0.30                        & 85.30           & 83.68          & 86.47          & 84.78          & 83.00              & 81.59          & 84.92                      & 83.35                    \\
                          & 0.40                        & 85.43          & 83.86          & 86.60           & 84.96          & 83.07           & 81.70           & 85.03                      & 83.51                    \\
                          & 0.50                        & 85.67          & 84.18          & 86.90           & 85.34          & 83.27           & 81.99          & 85.28                      & 83.84                    \\
                          & 0.60                        & 86.13          & 84.78          & 87.37          & 85.95          & 83.57           & 82.42          & 85.69                      & 84.38                    \\
                          & 0.70                        & 86.83          & 85.76          & 88.23          & 87.07          & \textbf{84.07}  & 83.22          & 86.38                      & 85.35                    \\
                          & 0.80                        & 86.93          & 86.20           & 89.13          & 88.25          & 83.50            & 83.18          & 86.52                      & 85.88                    \\
                          & 0.90                        & \textbf{87.50}  & \textbf{87.10}  & \textbf{89.80}  & \textbf{89.19} & 83.13           & \textbf{83.33} & \textbf{86.81}             & \textbf{86.54}           \\ \hline
\end{tabular}%
}
\caption{Hyperparameter tunning for $\alpha$ and $\lambda$.}
\label{tab: hyperparam}
\end{table}

\begin{table}[htbp]
\small
\centering
\resizebox{\columnwidth}{!}{%
\begin{tabular}{|ccccccccc|}
\hline
\rowcolor[HTML]{C0C0C0} 
\multicolumn{9}{|c|}{\cellcolor[HTML]{C0C0C0}POPE MSCOCO}                                                                                                                                                                                                \\ \hline
\multicolumn{1}{|c|}{}                               & \multicolumn{2}{c}{Popular}     & \multicolumn{2}{c}{Random}      & \multicolumn{2}{c|}{Adversarial}                                    &                             &                           \\ \cline{2-7}
\multicolumn{1}{|c|}{\multirow{-2}{*}{Method}}       & Acc.           & F1             & Acc.           & F1             & Acc.           & \multicolumn{1}{c|}{F1}                            & \multirow{-2}{*}{Acc.(avg)} & \multirow{-2}{*}{F1(avg)} \\ \hline
\multicolumn{1}{|c|}{Greedy}                         & 85.80           & 84.29          & 87.03          & 85.46          & 83.50           & \multicolumn{1}{c|}{82.19}                         & 85.44                       & \textbf{83.98}            \\
\multicolumn{1}{|c|}{w/ VCD}                         & 85.57          & 84.10           & 86.87          & 85.32          & 83.10           & \multicolumn{1}{c|}{81.87}                         & 85.18                       & 83.76                     \\
\multicolumn{1}{|c|}{w/ DoLa}                        & \textbf{85.90}  & \textbf{84.41} & \textbf{87.13} & \textbf{85.58} & \textbf{83.63} & \multicolumn{1}{c|}{\textbf{82.33}}                & \textbf{85.55}              & 84.11                     \\
\rowcolor[HTML]{C6F1C6} 
\multicolumn{1}{|c|}{\cellcolor[HTML]{C6F1C6}w/ TPC} & 85.73          & 84.21          & 87.00             & 85.40           & 83.50           & \multicolumn{1}{c|}{\cellcolor[HTML]{C6F1C6}82.16} & 85.41                       & 83.92                     \\ \hline
\rowcolor[HTML]{C0C0C0} 
\multicolumn{9}{|c|}{\cellcolor[HTML]{C0C0C0}POPE A-OKVQA}                                                                                                                                                                                               \\ \hline
\multicolumn{1}{|c|}{}                               & \multicolumn{2}{c}{Popular}     & \multicolumn{2}{c}{Random}      & \multicolumn{2}{c|}{Adversarial}                                    &                             &                           \\ \cline{2-7}
\multicolumn{1}{|c|}{\multirow{-2}{*}{Method}}       & Acc.           & F1             & Acc.           & F1             & Acc.           & \multicolumn{1}{c|}{F1}                            & \multirow{-2}{*}{Acc.(avg)} & \multirow{-2}{*}{F1(avg)} \\ \hline
\multicolumn{1}{|c|}{Greedy}                         & 85.27          & 85.05          & 88.70           & 88.12          & 78.83          & \multicolumn{1}{c|}{79.83}                         & 84.27                       & 84.33                     \\
\multicolumn{1}{|c|}{w/ VCD}                         & 85.00             & 84.91          & \textbf{88.80}  & \textbf{88.28} & 78.53          & \multicolumn{1}{c|}{79.68}                         & 84.11                       & 84.29                     \\
\multicolumn{1}{|c|}{w/ DoLa}                        & 85.30           & 85.10           & 88.73          & 88.17          & \textbf{78.90}  & \multicolumn{1}{c|}{\textbf{79.91}}                & 84.31                       & 84.39                     \\
\rowcolor[HTML]{C6F1C6} 
\multicolumn{1}{|c|}{\cellcolor[HTML]{C6F1C6}w/ TPC} & \textbf{85.37} & \textbf{85.14} & 88.73          & 88.16          & \textbf{78.90}  & \multicolumn{1}{c|}{\cellcolor[HTML]{C6F1C6}79.90}  & \textbf{84.33}              & \textbf{84.40}            \\ \hline
\rowcolor[HTML]{C0C0C0} 
\multicolumn{9}{|c|}{\cellcolor[HTML]{C0C0C0}POPE GQA}                                                                                                                                                                                                   \\ \hline
\multicolumn{1}{|c|}{}                               & \multicolumn{2}{c}{Popular}     & \multicolumn{2}{c}{Random}      & \multicolumn{2}{c|}{Adversarial}                                    &                             &                           \\ \cline{2-7}
\multicolumn{1}{|c|}{\multirow{-2}{*}{Method}}       & Acc.           & F1             & Acc.           & F1             & Acc.           & \multicolumn{1}{c|}{F1}                            & \multirow{-2}{*}{Acc.(avg)} & \multirow{-2}{*}{F1(avg)} \\ \hline
\multicolumn{1}{|c|}{Greedy}                         & 83.97          & 84.09          & 89.33          & 88.82          & 80.83          & \multicolumn{1}{c|}{81.55}                         & 84.71                       & 84.82                     \\
\multicolumn{1}{|c|}{w/ VCD}                         & 83.83          & 84.06          & 89.27          & 88.82          & 80.80           & \multicolumn{1}{c|}{81.64}                         & 84.63                       & 84.84                     \\
\multicolumn{1}{|c|}{w/ DoLa}                        & \textbf{84.03} & \textbf{84.15} & \textbf{89.37} & \textbf{88.86} & \textbf{80.90}  & \multicolumn{1}{c|}{\textbf{81.62}}                & \textbf{84.77}              & \textbf{84.88}            \\
\rowcolor[HTML]{C6F1C6} 
\multicolumn{1}{|c|}{\cellcolor[HTML]{C6F1C6}w/ TPC} & 83.87          & 83.96          & \textbf{89.37} & 88.82          & 80.77          & \multicolumn{1}{c|}{\cellcolor[HTML]{C6F1C6}81.45} & 84.67                       & 84.74                     \\ \hline
\end{tabular}%
}
\caption{Results of TPC and baselines in greedy search.}
\label{tab: Greedy}
\end{table}

\begin{figure*}[htbp]
    \centering
    \begin{subfigure}{\columnwidth}
        \centering
        \includegraphics[width=\columnwidth]{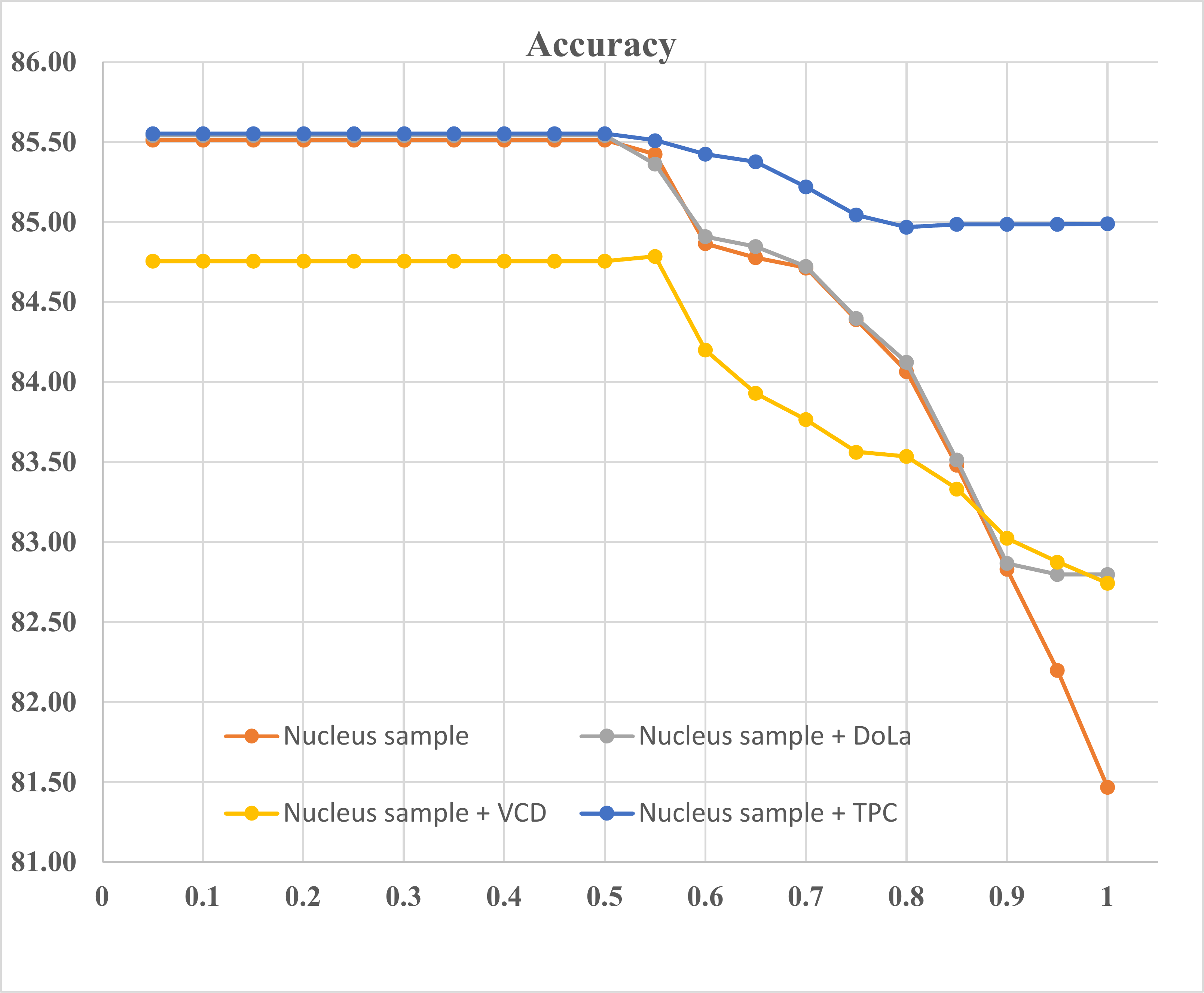}
    \end{subfigure}
    \begin{subfigure}{\columnwidth}
        \centering
        \includegraphics[width=\columnwidth]{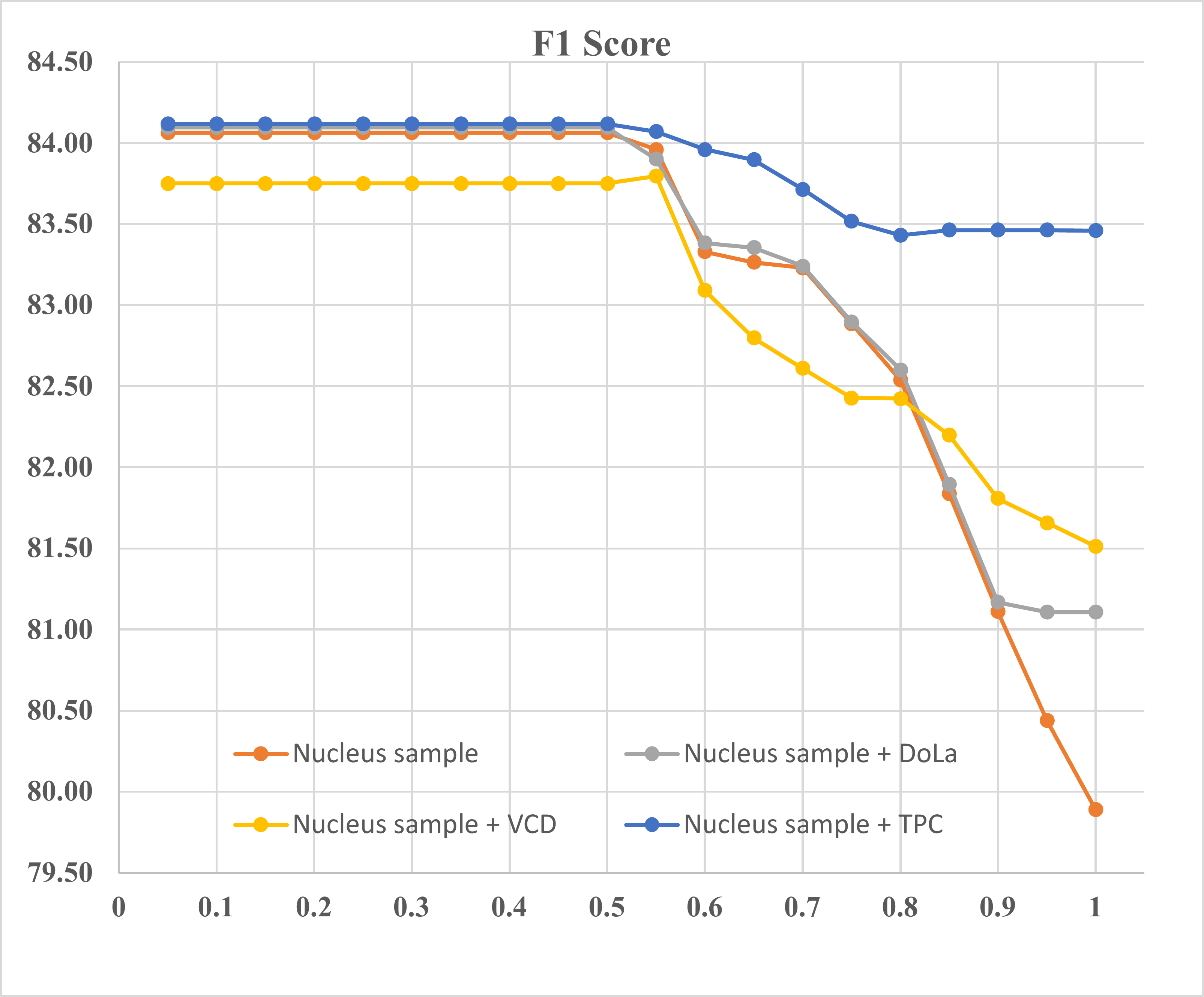}
    \end{subfigure}
    \caption{Baselines and TPC with Nucleus Sampling, the x-axis represents the top-p value.}
    \label{fig:Nucleus}
\end{figure*}

\begin{table*}[]
\centering
\resizebox{\linewidth}{!}{%
\begin{tabular}{|ccccccccccccccccccccc|}
\hline
\multicolumn{21}{|c|}{\cellcolor[HTML]{C0C0C0}POPE MSCOCO}                                                                                                                                                                                                                                                                                                                                                                                                                                                                                                                                                                                 \\ \hline
\multicolumn{1}{|c|}{}                           & \multicolumn{1}{c|}{}                        & \multicolumn{2}{c}{Popular}     & \multicolumn{2}{c}{Random}      & \multicolumn{2}{c|}{Adversarial}                     &                             & \multicolumn{1}{c|}{}                          & \multicolumn{1}{c|}{}                    & \multicolumn{1}{c|}{}                           & \multicolumn{1}{c|}{}                        & \multicolumn{2}{c}{Popular}     & \multicolumn{2}{c}{Random}      & \multicolumn{2}{c|}{Adversarial}                     &                             &                           \\ \cline{3-8} \cline{14-19}
\multicolumn{1}{|c|}{\multirow{-2}{*}{Method}}   & \multicolumn{1}{c|}{\multirow{-2}{*}{Top-p}} & Acc.           & F1             & Acc.           & F1             & Acc.           & \multicolumn{1}{c|}{F1}             & \multirow{-2}{*}{Acc.(avg)} & \multicolumn{1}{c|}{\multirow{-2}{*}{F1(avg)}} & \multicolumn{1}{c|}{}                    & \multicolumn{1}{c|}{\multirow{-2}{*}{Method}}   & \multicolumn{1}{c|}{\multirow{-2}{*}{Top-p}} & Acc.           & F1             & Acc.           & F1             & Acc.           & \multicolumn{1}{c|}{F1}             & \multirow{-2}{*}{Acc.(avg)} & \multirow{-2}{*}{F1(avg)} \\ \cline{1-10} \cline{12-21} 
\multicolumn{1}{|c|}{}                           & \multicolumn{1}{c|}{0.50}                     & \textbf{85.87} & \textbf{84.38} & \textbf{87.10}  & \textbf{85.54} & \textbf{83.57} & \multicolumn{1}{c|}{\textbf{82.27}} & \textbf{85.51}              & \multicolumn{1}{c|}{\textbf{84.06}}            & \multicolumn{1}{c|}{}                    & \multicolumn{1}{c|}{}                           & \multicolumn{1}{c|}{0.50}                     & \textbf{85.07} & \textbf{84.00}    & \textbf{86.90}  & \textbf{85.68} & 82.30           & \multicolumn{1}{c|}{81.57}          & 84.76                       & 83.75                     \\
\multicolumn{1}{|c|}{}                           & \multicolumn{1}{c|}{0.55}                    & 85.87          & 84.35          & 86.90           & 85.33          & 83.50           & \multicolumn{1}{c|}{82.20}           & 85.42                       & \multicolumn{1}{c|}{83.96}                     & \multicolumn{1}{c|}{}                    & \multicolumn{1}{c|}{}                           & \multicolumn{1}{c|}{0.55}                    & 85.03          & 83.99          & 86.80           & 85.61          & \textbf{82.53} & \multicolumn{1}{c|}{\textbf{81.79}} & \textbf{84.79}              & \textbf{83.80}            \\
\multicolumn{1}{|c|}{}                           & \multicolumn{1}{c|}{0.6}                     & 85.30           & 83.71          & 86.50           & 84.84          & 82.80           & \multicolumn{1}{c|}{81.45}          & 84.87                       & \multicolumn{1}{c|}{83.33}                     & \multicolumn{1}{c|}{}                    & \multicolumn{1}{c|}{}                           & \multicolumn{1}{c|}{0.60}                     & 84.60           & 83.42          & 86.13          & 84.82          & 81.87          & \multicolumn{1}{c|}{81.03}          & 84.20                       & 83.09                     \\
\multicolumn{1}{|c|}{}                           & \multicolumn{1}{c|}{0.65}                    & 85.20           & 83.64          & 86.53          & 84.89          & 82.60           & \multicolumn{1}{c|}{81.26}          & 84.78                       & \multicolumn{1}{c|}{83.26}                     & \multicolumn{1}{c|}{}                    & \multicolumn{1}{c|}{}                           & \multicolumn{1}{c|}{0.65}                    & 84.23          & 83.03          & 86.13          & 84.76          & 81.43          & \multicolumn{1}{c|}{80.60}           & 83.93                       & 82.80                     \\
\multicolumn{1}{|c|}{}                           & \multicolumn{1}{c|}{0.70}                     & 85.07          & 83.54          & 86.57          & 84.95          & 82.50           & \multicolumn{1}{c|}{81.20}           & 84.71                       & \multicolumn{1}{c|}{83.23}                     & \multicolumn{1}{c|}{}                    & \multicolumn{1}{c|}{}                           & \multicolumn{1}{c|}{0.70}                     & 84.13          & 82.90           & 86.00             & 84.60           & 81.17          & \multicolumn{1}{c|}{80.33}          & 83.77                       & 82.61                     \\
\multicolumn{1}{|c|}{}                           & \multicolumn{1}{c|}{0.75}                    & 84.87          & 83.31          & 86.20           & 84.55          & 82.10           & \multicolumn{1}{c|}{80.80}           & 84.39                       & \multicolumn{1}{c|}{82.89}                     & \multicolumn{1}{c|}{}                    & \multicolumn{1}{c|}{}                           & \multicolumn{1}{c|}{0.75}                    & 83.93          & 82.71          & 85.93          & 84.53          & 80.83          & \multicolumn{1}{c|}{80.04}          & 83.56                       & 82.43                     \\
\multicolumn{1}{|c|}{}                           & \multicolumn{1}{c|}{0.80}                     & 84.5           & 82.91          & 85.93          & 84.24          & 81.77          & \multicolumn{1}{c|}{80.46}          & 84.07                       & \multicolumn{1}{c|}{82.54}                     & \multicolumn{1}{c|}{}                    & \multicolumn{1}{c|}{}                           & \multicolumn{1}{c|}{0.80}                     & 83.97          & 82.77          & 85.97          & 84.58          & 80.67          & \multicolumn{1}{c|}{79.92}          & 83.54                       & 82.42                     \\
\multicolumn{1}{|c|}{}                           & \multicolumn{1}{c|}{0.85}                    & 83.97          & 82.26          & 85.40           & 83.58          & 81.07          & \multicolumn{1}{c|}{79.67}          & 83.48                       & \multicolumn{1}{c|}{81.84}                     & \multicolumn{1}{c|}{}                    & \multicolumn{1}{c|}{}                           & \multicolumn{1}{c|}{0.85}                    & 83.70           & 82.48          & 85.70           & 84.29          & 80.60           & \multicolumn{1}{c|}{79.82}          & 83.33                       & 82.20                     \\
\multicolumn{1}{|c|}{}                           & \multicolumn{1}{c|}{0.90}                     & 83.43          & 81.63          & 84.63          & 82.73          & 80.43          & \multicolumn{1}{c|}{78.97}          & 82.83                       & \multicolumn{1}{c|}{81.11}                     & \multicolumn{1}{c|}{}                    & \multicolumn{1}{c|}{}                           & \multicolumn{1}{c|}{0.90}                     & 83.27          & 81.98          & 85.47          & 83.97          & 80.33          & \multicolumn{1}{c|}{79.47}          & 83.02                       & 81.81                     \\
\multicolumn{1}{|c|}{}                           & \multicolumn{1}{c|}{0.95}                    & 82.87          & 81.01          & 84.03          & 82.07          & 79.70           & \multicolumn{1}{c|}{78.23}          & 82.20                        & \multicolumn{1}{c|}{80.44}                     & \multicolumn{1}{c|}{}                    & \multicolumn{1}{c|}{}                           & \multicolumn{1}{c|}{0.95}                    & 83.27          & 81.96          & 85.23          & 83.73          & 80.13          & \multicolumn{1}{c|}{79.28}          & 82.88                       & 81.66                     \\
\multicolumn{1}{|c|}{\multirow{-11}{*}{Nucleus}} & \multicolumn{1}{c|}{1.00}                       & 82.10           & 80.28          & 83.30           & 81.35          & 79.00             & \multicolumn{1}{c|}{78.03}          & 81.47                       & \multicolumn{1}{c|}{79.89}                     & \multicolumn{1}{c|}{}                    & \multicolumn{1}{c|}{\multirow{-11}{*}{w/ VCD}}  & \multicolumn{1}{c|}{1.00}                       & 83.13          & 81.81          & 85.13          & 83.61          & 79.97          & \multicolumn{1}{c|}{79.11}          & 82.74                       & 81.51                     \\ \cline{1-10} \cline{12-21} 
\multicolumn{1}{|c|}{}                           & \multicolumn{1}{c|}{0.50}                     & \textbf{85.90}  & \textbf{84.42} & \textbf{87.13} & \textbf{85.59} & \textbf{83.63} & \multicolumn{1}{c|}{\textbf{82.34}} & \textbf{85.55}              & \multicolumn{1}{c|}{\textbf{84.12}}            & \multicolumn{1}{c|}{}                    & \multicolumn{1}{c|}{}                           & \multicolumn{1}{c|}{0.50}                     & \textbf{85.90}  & \textbf{84.41} & \textbf{87.13} & \textbf{85.58} & \textbf{83.60}  & \multicolumn{1}{c|}{\textbf{82.30}}  & \textbf{85.54}              & \textbf{84.10}            \\
\multicolumn{1}{|c|}{}                           & \multicolumn{1}{c|}{0.55}                    & 85.93          & 84.44          & 87.03          & 85.48          & 83.57          & \multicolumn{1}{c|}{82.29}          & 85.51                       & \multicolumn{1}{c|}{84.07}                     & \multicolumn{1}{c|}{}                    & \multicolumn{1}{c|}{}                           & \multicolumn{1}{c|}{0.55}                    & 85.83          & 84.31          & 86.83          & 85.26          & 83.43          & \multicolumn{1}{c|}{82.13}          & 85.36                       & 83.90                     \\
\multicolumn{1}{|c|}{}                           & \multicolumn{1}{c|}{0.6}                     & 85.87          & 84.35          & 86.90           & 85.33          & 83.50           & \multicolumn{1}{c|}{82.20}           & 85.42                       & \multicolumn{1}{c|}{83.96}                     & \multicolumn{1}{c|}{}                    & \multicolumn{1}{c|}{}                           & \multicolumn{1}{c|}{0.60}                     & 85.33          & 83.75          & 86.57          & 84.91          & 82.83          & \multicolumn{1}{c|}{81.49}          & 84.91                       & 83.38                     \\
\multicolumn{1}{|c|}{}                           & \multicolumn{1}{c|}{0.65}                    & 85.83          & 84.30           & 86.83          & 85.24          & 83.47          & \multicolumn{1}{c|}{82.15}          & 85.38                       & \multicolumn{1}{c|}{83.90}                     & \multicolumn{1}{c|}{}                    & \multicolumn{1}{c|}{}                           & \multicolumn{1}{c|}{0.65}                    & 85.27          & 83.73          & 86.60           & 84.98          & 82.67          & \multicolumn{1}{c|}{81.35}          & 84.85                       & 83.35                     \\
\multicolumn{1}{|c|}{}                           & \multicolumn{1}{c|}{0.7}                     & 85.63          & 84.08          & 86.70           & 85.08          & 83.33          & \multicolumn{1}{c|}{81.98}          & 85.22                       & \multicolumn{1}{c|}{83.71}                     & \multicolumn{1}{c|}{}                    & \multicolumn{1}{c|}{}                           & \multicolumn{1}{c|}{0.70}                     & 85.07          & 83.54          & 86.60           & 84.98          & 82.50           & \multicolumn{1}{c|}{81.20}           & 84.72                       & 83.24                     \\
\multicolumn{1}{|c|}{}                           & \multicolumn{1}{c|}{0.75}                    & 85.43          & 83.86          & 86.60           & 84.96          & 83.10           & \multicolumn{1}{c|}{81.73}          & 85.04                       & \multicolumn{1}{c|}{83.52}                     & \multicolumn{1}{c|}{}                    & \multicolumn{1}{c|}{}                           & \multicolumn{1}{c|}{0.75}                    & 84.90           & 83.34          & 86.20           & 84.55          & 82.10           & \multicolumn{1}{c|}{80.80}           & 84.40                       & 82.90                     \\
\multicolumn{1}{|c|}{}                           & \multicolumn{1}{c|}{0.8}                     & 85.37          & 83.78          & 86.57          & 84.91          & 82.97          & \multicolumn{1}{c|}{81.60}           & 84.97                       & \multicolumn{1}{c|}{83.43}                     & \multicolumn{1}{c|}{}                    & \multicolumn{1}{c|}{}                           & \multicolumn{1}{c|}{0.80}                     & 84.57          & 82.98          & 85.97          & 84.29          & 81.83          & \multicolumn{1}{c|}{80.53}          & 84.12                       & 82.60                     \\
\multicolumn{1}{|c|}{}                           & \multicolumn{1}{c|}{0.85}                    & 85.40           & 83.83          & 86.63          & 84.99          & 82.93          & \multicolumn{1}{c|}{81.57}          & 84.99                       & \multicolumn{1}{c|}{83.46}                     & \multicolumn{1}{c|}{}                    & \multicolumn{1}{c|}{}                           & \multicolumn{1}{c|}{0.85}                    & 84.00             & 82.31          & 85.47          & 83.67          & 81.07          & \multicolumn{1}{c|}{79.70}           & 83.51                       & 81.89                     \\
\multicolumn{1}{|c|}{}                           & \multicolumn{1}{c|}{0.90}                     & 85.40           & 83.83          & 86.63          & 84.99          & 82.93          & \multicolumn{1}{c|}{81.57}          & 84.99                       & \multicolumn{1}{c|}{83.46}                     & \multicolumn{1}{c|}{}                    & \multicolumn{1}{c|}{}                           & \multicolumn{1}{c|}{0.90}                     & 83.47          & 81.68          & 84.70           & 82.82          & 80.43          & \multicolumn{1}{c|}{79.00}             & 82.87                       & 81.17                     \\
\multicolumn{1}{|c|}{}                           & \multicolumn{1}{c|}{0.95}                    & 85.40           & 83.83          & 86.63          & 84.99          & 82.93          & \multicolumn{1}{c|}{81.57}          & 84.99                       & \multicolumn{1}{c|}{83.46}                     & \multicolumn{1}{c|}{}                    & \multicolumn{1}{c|}{}                           & \multicolumn{1}{c|}{0.95}                    & 83.43          & 81.65          & 84.67          & 82.78          & 80.30           & \multicolumn{1}{c|}{78.89}          & 82.80                       & 81.11                     \\
\multicolumn{1}{|c|}{\multirow{-11}{*}{w/ TPC}}  & \multicolumn{1}{c|}{1.00}                       & 85.40           & 83.83          & 86.63          & 84.99          & 82.93          & \multicolumn{1}{c|}{81.57}          & 84.99                       & \multicolumn{1}{c|}{83.46}                     & \multicolumn{1}{c|}{\multirow{-24}{*}{}} & \multicolumn{1}{c|}{\multirow{-11}{*}{w/ DoLa}} & \multicolumn{1}{c|}{1.00}                       & 83.43          & 81.65          & 84.67          & 82.78          & 80.30           & \multicolumn{1}{c|}{78.89}          & 82.80                        & 81.11                     \\ \hline
\end{tabular}%
}
\caption{\textbf{Nucleus sampling}. Adjust the top-p value and evaluate the effect of baselines’ performance on POPE MSCOCO.}
\label{tab: Nucleus}
\end{table*}

\section{Efficiency}
\label{sec: efficiency}

\begin{table}[H]
\centering
\resizebox{\columnwidth}{!}{%
\begin{tabular}{|c|c|c|c|}
\hline
Method  & Throughput (samples/s) $\uparrow$ & Latency (ms/sample) $\downarrow$ & GPU Memory (MiB) $\downarrow$ \\ \hline
Regular & \textbf{4.03}                     & \textbf{248}                     & \textbf{15391}                \\
w/ VCD  & 2.42                              & 413                              & 16031                         \\
w/ DoLa & 3.94                              & 254                              & 16145                         \\
w/ TPC  & 3.97                              & 252                              & \textbf{15391}                \\ \hline
\end{tabular}%
}
\caption{Comparison of throughput, latency, and memory usage.}
\label{efficiency}
\end{table}

Since TPC does not require comparisons with any other logits and does not involve regenerating logits at the same time step, it greatly reduces additional computational overhead. In Table \ref{efficiency}, we compare the throughput and memory usage of Regular, VCD, and DoLa. The results indicate that TPC is an almost lossless post-hoc method. VCD nearly doubles inference time compared to Regular, while DoLa, although not reducing inference speed, requires additional memory due to the need to search for premature layers.

\section{ Greedy Search and Nucleus Sampling}
\label{sec: greedy nucleus}
\subsection{Greedy Search}
Table \ref{tab: Greedy} presents the performance of greedy search on POPE MSCOCO. Greedy search is applied to both baselines and TPC, and the results show minimal differences in performance across all methods. Since greedy search selects the highest probability token at each timestep, these post-hoc methods have limited influence on adjusting the value of the most probable token, leading to only minor variations. Consequently, these experimental findings suggest that VCD, DoLa, and TPC are more effective when used with sampling method, helping to achieve a better balance between diversity and accuracy in text generation.

\subsection{Nucleus Sampling}
In Table \ref{tab: Nucleus}, we combine nucleus sampling with each post-hoc method. We gradually increase the top-p value from 0.05 to 1.00, where top-p = 1.00 corresponds to direct sampling, also referred to as Regular above. The curves in Figure \ref{fig:Nucleus} reveal that both the baselines and TPC maintain stable F1 scores and accuracy when top-p is within the range of 0.05 to 0.5. As top-p increases further, performance begins to decline across all methods. The baselines experience varying degrees of degradation, while TPC shows only a slight drop, which quickly stabilizes with minimal decline.

\section{Cases}
\label{sec: cases}
\begin{figure*}[htbp]
    \centering
    \includegraphics[width=\linewidth]{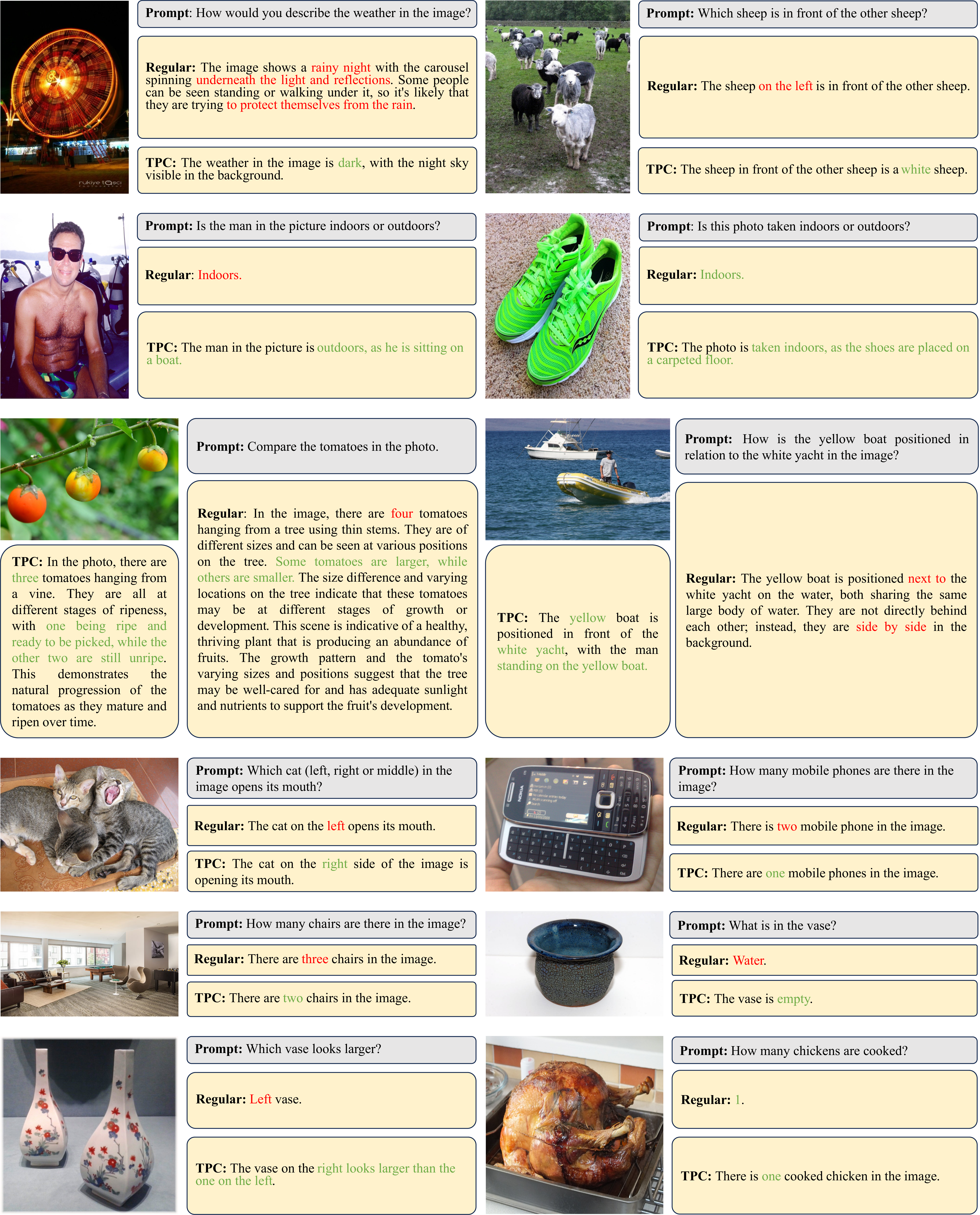}
    \caption{More cases from MMHal-Bench.}
    \label{fig:add cases}
\end{figure*}

In Figure \ref{fig:add cases}, we select additional open-ended text generated by Regular and TPC for comparison on MMHal-Bench. When answering object-level and attribute-level questions, Regular often provides correct responses but tends to generate only a single token. Conversely, it frequently produces hallucinated tokens when generating detailed image descriptions. In contrast, TPC generates responses that are more accurate and balanced.
\newpage

\end{document}